\documentclass[twoside]{article}

\usepackage[accepted]{aistats2024}
\usepackage[utf8]{inputenc} % allow utf-8 input
\usepackage[T1]{fontenc}    % use 8-bit T1 fonts
\usepackage{hyperref}       % hyperlinks
\usepackage{url}            % simple URL typesetting
\usepackage{booktabs}       % professional-quality tables
\usepackage{amsfonts}       % blackboard math symbols
\usepackage{nicefrac}       % compact symbols for 1/2, etc.
\usepackage{microtype}      % microtypography
\usepackage{xcolor}      
\usepackage[round]{natbib}
\usepackage{amsthm}
\usepackage{amsmath}
\usepackage{soul}
\theoremstyle{plain}
\newtheorem{theorem}{Theorem}

\newtheorem{lemma}{Lemma}
\newtheorem{corollary}{Corollary}
\theoremstyle{definition}
\newtheorem{definition}{Definition}

\theoremstyle{plain}

\usepackage{wrapfig}
\usepackage{graphicx}
\usepackage{subfigure}
\usepackage{enumitem}
\usepackage{selectp}
\newcommand\given[1][]{\:#1\vert\:}

\usepackage{xr}
\makeatletter
\newcommand*{\addFileDependency}[1]{% argument=file name and extension
  \typeout{(#1)}
  \@addtofilelist{#1}
  \IfFileExists{#1}{}{\typeout{No file #1.}}
}
\makeatother

\newcommand*{\myexternaldocument}[1]{%
    \externaldocument{#1}%
    \addFileDependency{#1.tex}%
    \addFileDependency{#1.aux}%
}
\myexternaldocument{supplement}

\begin{document}

\runningtitle{Analyzing Explainer Robustness via Probabilistic Lipschitzness of Prediction Functions}

\runningauthor{Zulqarnain Khan$^*$, Davin Hill$^*$, Aria Masoomi, Josh Bone, Jennifer Dy}

\twocolumn[

\aistatstitle{Analyzing Explainer Robustness via \\ Probabilistic Lipschitzness of Prediction Functions}

\aistatsauthor{ Zulqarnain Khan$^*$ \And Davin Hill$^*$ \And Aria Masoomi}
\aistatsaddress{Northeastern University \\ khanzu@ece.neu.edu \And Northeastern University \\ dhill@ece.neu.edu   \And Northeastern University \\ masoomi.a@northeastern.edu}

% \aistatsauthor{ Author 1 \And Author 2 \And  Author 3 }

% \aistatsaddress{ Institution 1 \And  Institution 2 \And Institution 3 } ]

\aistatsauthor{ Josh Bone \And Jennifer Dy}
\aistatsaddress{Northeastern University \\ bone.j@northeastern.edu\And Northeastern University \\ jdy@ece.neu.edu} ]

% \aistatsauthor{Zulqarnain Khan, Davin Hill, Aria Masoomi, et al}
% \aistatsaddress{Department of Electrical and Computer Engineering \\ Northeastern University \\ khanzu@ece.neu.edu \\ }

\begin{abstract}
Machine learning methods have significantly improved in their predictive capabilities, but at the same time they are becoming more complex and less transparent. As a result, explainers are often relied on to provide interpretability to these \textit{black-box} prediction models. As crucial diagnostics tools, it is important that these explainers themselves are robust. In this paper we focus on one particular aspect of robustness, namely that an explainer should give similar explanations for similar data inputs. We formalize this notion by introducing and defining \textit{explainer astuteness}, analogous to astuteness of prediction functions. Our formalism allows us to connect explainer robustness to the predictor's \emph{probabilistic Lipschitzness}, which captures the probability of local smoothness of a function. We provide lower bound guarantees on the astuteness of a variety of explainers (e.g., SHAP, RISE, CXPlain) given the Lipschitzness of the prediction function. These theoretical results imply that locally smooth prediction functions lend themselves to locally robust explanations. We evaluate these results empirically on simulated as well as real datasets.
\end{abstract}

\section{INTRODUCTION} \label{sec:intro}

Machine learning models have made significant improvements in their ability to predict and classify data. However, these gains in predictive power have come at the cost of increasingly opaque models, which has resulted in a proliferation of explainers that seek to provide transparency for these black-box models. Given the importance of these explainers, it is essential to understand the factors that contribute to their robustness and effectiveness.

In this paper we focus on explainer robustness. A robust explainer is one where \emph{similar inputs results in similar explanations}. As an example, consider two patients given the same diagnosis in a medical setting. These two patients share identical symptoms and are demographically very similar, therefore a clinician would expect that factors influencing the model decision should be similar as well. Prior work in explainer robustness suggests that this expectation does not always hold true \citep{alvarez2018robustness, ghorbani2019interpretation}; small changes to the input samples can result in large shifts in explanation.

\begin{figure}[t]
    \centering
    \includegraphics[width=\columnwidth]{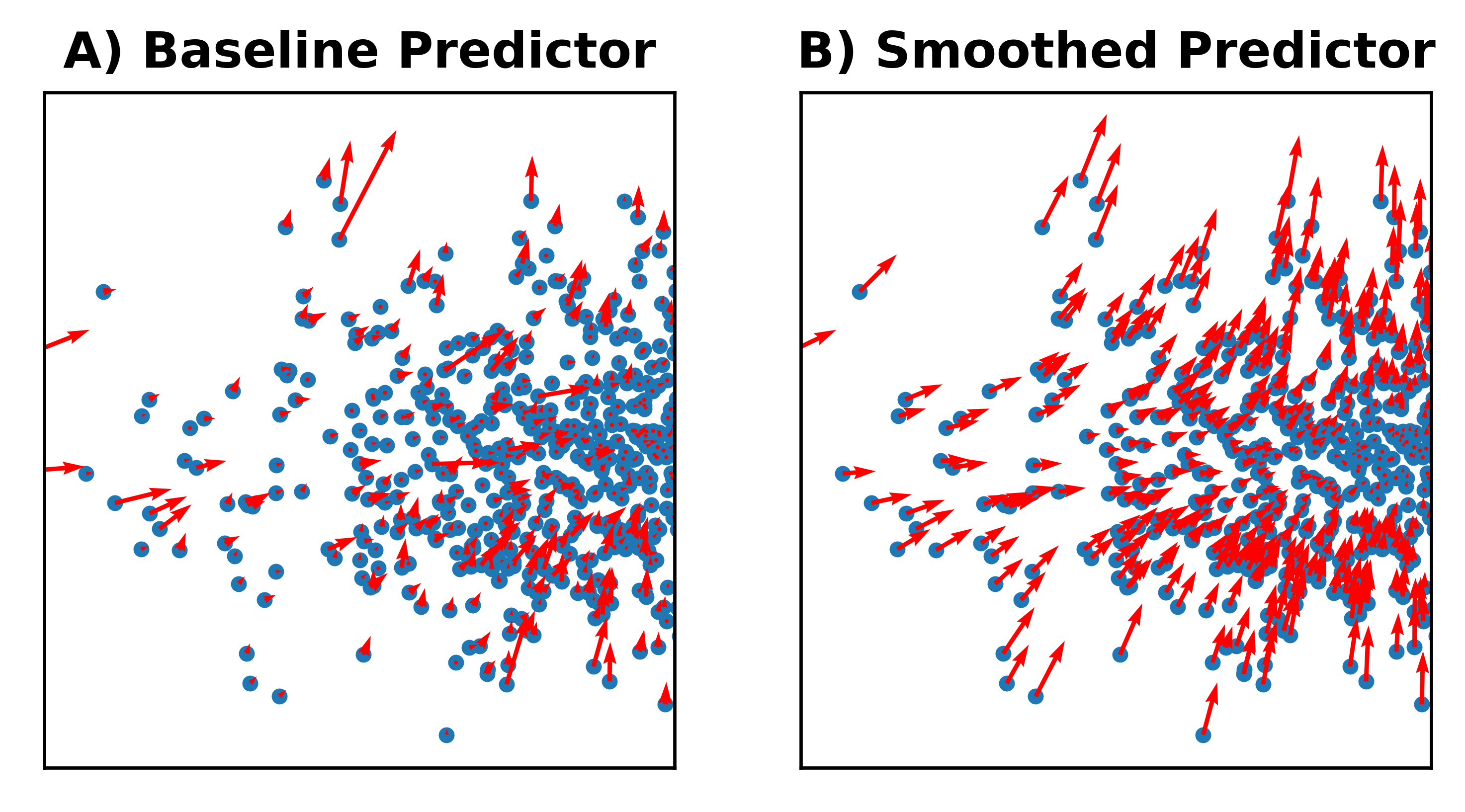}
    \setlength{\abovecaptionskip}{-5pt}
    \caption{\textbf{Smoother black-box predictors lead to more astute explainers and robust explanations.} Samples from a simulated dataset are plotted in blue, with red arrows representing respective post-hoc SHAP explanations plotted as a vector. \textbf{A)} When a neural network (NN) is trained with no Lipschitz constraints, explanations of nearby points can vary significantly, as evidenced by the arrows varying in length and direction. \textbf{B)} When the NN is retrained with Lipschitz regularization suggested by \cite{gouk2021regularisation}, explanations are observed to be more aligned in length and direction, indicating higher robustness.}
    \label{fig:intro_toy}
\end{figure}

For this reason we investigate the theoretical underpinning of explainer robustness. Specifically, we focus on investigating the connection between explainer robustness and smoothness of the black-box model being explained.
% What do we do in this paper?
In this work, we propose and formally define \emph{Explainer Astuteness}, which characterizes the ability of a given explainer to provide robust explanations.
% -- a property of explainers which captures the probability that a given method provides similar explanations to similar data points. 
% Explainer Astuteness allows us to evaluate the robustness for a given explainer over the entire dataset and helps tie explainer robustness to probabilistic Lipschitzness of classifiers.
Explainer Astuteness enables the evaluation of different explainers for prediction tasks where having robust explanations is critical.
We then establish a theoretical connection between explainer astuteness and the \emph{probabilistic Lipschitzness} of the black-box model that is being explained. 
Since probabilistic Lipschitzness is a measure of the probability that a function is smooth in a local neighborhood, our results demonstrate how the smoothness of the black-box model itself impacts the astuteness of the explainer (Fig. \ref{fig:intro_toy}).

\textbf{Contributions:}
\begin{itemize}[leftmargin=2em, topsep=0em, itemsep=0em]
    \item  We formalize \emph{explainer astuteness}, which captures how well a given explainer provides similar explanations to similar points. This formalism allows us to evaluate different explainers based on the robustness of their explanations.
    
    \item We establish a lower bound on explainer astuteness which is dependent on the smoothness of the black-box model. We characterize model smoothness with \emph{probabilistic Lipschitzness}, which allows for a broader class of black-box models than with standard Lipschitzness. 
    % \item We provide theoretical results that connect astuteness of explainers to the smoothness of the black-box model they are providing explanations on. \textbf{Our results suggest that smooth black-box models result in explainers providing more astute explanations}. While this statement is intuitive, proving it is non-trivial and requires additional assumptions for different explainers (See Section~\ref{subsec:impl}).
    
    \item We investigate the astuteness of three classes of explainers: (1) Shapley value-based (e.g., SHAP), (2) explainers that simulate mean effect of features (e.g., RISE), and (3) explainers that simulate individual feature removal (e.g., CXPlain).
    
    \item Our theoretical results are empirically validated on a mix of simulated and real datasets. Experiments show that increasing the smoothness of black-box models also improves the astuteness of the three classes of investigated explainers.
    % \item We demonstrate experimentally that this lower bound indeed holds in practice by comparing the astuteness predicted by our theorems to the observed astuteness on simulated and real datasets. We also demonstrate experimentally that the same neural network when trained with Lipschitz constraints lends itself to more astute explanations compared to when it is trained with no constraints.
\end{itemize}

\section{RELATED WORKS} 
\vspace{-2mm}
% Explainers
\textbf{Explainers.} A wide variety of explainers have been proposed in the literature \citep{guidotti2018survey,arrieta2020explainable}. We focus on \emph{post-hoc} methods, specifically feature attribution and feature selection explainers.
Feature attribution explainers provide continuous-valued importance scores to each of the input features.
Some models such as CXPlain \citep{schwab2019cxplain}, PredDiff \citep{zintgraf2017visualizing} and feature ablation explainers \citep{lei2018distribution} calculate feature attributions by simulating individual feature removal, while other methods such as RISE \citep{petsiuk2018rise} calculate the mean effect of a feature's presence to attribute importance to it. 
\cite{lundberg2017unified} unify six different feature attribution explainers under the SHAP framework.
Other works have extended feature attributions to higher-order explanations \citep{treeshap, masoomi2022explanations, torop2023smoothhess}.
In contrast, feature selection explainers provide binary explanations for each feature and include individual selector approaches such as L2X \citep{chen2018learning} and INVASE \citep{yoon2018invase}, and group-wise selection approaches such as gI \citep{masoomi2020instance}.

In this work, we focus on \emph{removal-based explainers}.
% \cite{covert2020explaining} combined 25 existing methods under the class of \textit{removal-based explainers}.
Removal based feature explainers are a class of 25 methods defined by \cite{covert2020explaining} that define a feature's influence through the impact of removing it from a model. This includes popular approaches including KernelSHAP, LIME, DeepLIFT \citep{lundberg2017unified}, mean effect based methods such as RISE \citep{petsiuk2018rise}, and individual effects based methods such as CXPlain \citep{schwab2019cxplain}, PredDiff \citep{zintgraf2017visualizing}, permutation tests \citep{strobl2008conditional}, and feature ablation explainers \citep{lei2018distribution}. All of these methods simulate feature removal either explicitly or implicitly. For example, SHAP explicitly considers effect of using subsets that include a feature as compared to the effect of removing that feature from the subset.

% Robustness
\textbf{Explainer Robustness.}
Similarly, there has been a recent increase in research focused on analyzing different aspects of explainer behavior and reliability.
\cite{yin2021faithfulness} propose stability and sensitivity as measures of faithfulness of explainers to the classifier decision-making process.
\cite{yeh2019fidelity} investigate infidelity and sensitivity of explanations under perturbations.
\cite{li2020learning} explore connections between explainers and model generalization. 
\cite{agarwal2022probing} investigate theoretical guarantees for stability of Graph Neural Network explainers.
The term ``robustness'' for explainers has also been used in different contexts, such as in relation to distribution shifts \citep{lakkaraju2020robust, upadhyay2021towards}, model geometry \citep{dombrowski_explanations_can_be_manipulated, wang2020smoothed, hill2022explanation}, or adversarial attacks \citep{ghorbani2019interpretation, rieger2020simple}. We follow the definition by \cite{alvarez2018robustness}, who empirically show that robustness, in the sense that explainers should provide similar explanations for similar inputs, is a desirable property and how forcing this property yields better explanations. Recently, \cite{agarwal2021towards} explore the robustness of LIME \citep{ribeiro2016should} and SmoothGrad \citep{smilkov2017smoothgrad}, and prove that, for these two methods, their robustness is related to the maximum value of the gradient of the predictor function.

% How our work is different
Our work is related to the work by \cite{alvarez2018robustness} and \cite{agarwal2021towards} on explainer robustness. However, instead of enforcing explainers to be robust themselves \citep{alvarez2018robustness}, our theoretical results suggest that ensuring robustness of explanations also depends on the smoothness of the black-box model that is being explained. Our results are complementary to the results obtained by \cite{agarwal2021towards} in that our theorems cover a wider variety of explainers as compared to only  Continuous LIME and SmoothGrad.  We further relate robustness to probabilistic Lipschitzness of black-box models, which is a quantity that can be empirically estimated. 
Our work is also related to independently-developed contemporaneous work by \cite{tan2023robust} and \cite{agarwal2022rethinking}. However, our approach focuses on \emph{probabilistic} Lipschitzness and its relationship to astuteness, which can be likened to a probabilistic form of ``robustness'' over the entire data distribution. This allows us to address classifiers that fall on a spectrum between robust and non-robust, rather than adhering to deterministic binary categorizations as in \cite{tan2023robust}. In addition, our approach contrasts with that of \cite{agarwal2022rethinking}, which only considers \emph{local} robustness. 

\textbf{Lipschitzness of Neural Networks.} There has been recent work estimating upper-bounds of Lipschitz constant for neural networks \citep{virmaux2018lipschitz, fazlyab2019efficient, gouk2021regularisation}, and enforcing Lipschitz continuity  during neural networks training, with an eye towards improving classifier robustness \citep{gouk2021regularisation,aziznejad2020deep, fawzi2017robustness, alemi2016deep}. \citep{fel2022good} empirically demonstrated that 1-Lipschitz networks are better suited as predictors that are more explainable and trustworthy.
Other benefits of neural network smoothness have been explored \citep{pmlr-v137-rosca20a}, such as in improving model generalization \citep{hardt2016train, nakkiran2021deep} and adversarial robustness \citep{novak2018sensitivity}.
Our work provides crucial additional motivation for this line of research; i.e., it provides theoretical reasons to improve Lipschitzness of neural networks from the perspective of enabling more robust explanations by proving that \emph{enforcing smoothness on black-box models lends them to more robust explanations.}
%===============================================================================
%===============================================================================

\section{EXPLAINER ASTUTENESS} \label{sec:analysis}
% \vspace{-2mm}
    
\begin{figure*}[!t]
    \begin{center}
    \includegraphics[width=0.91\linewidth]{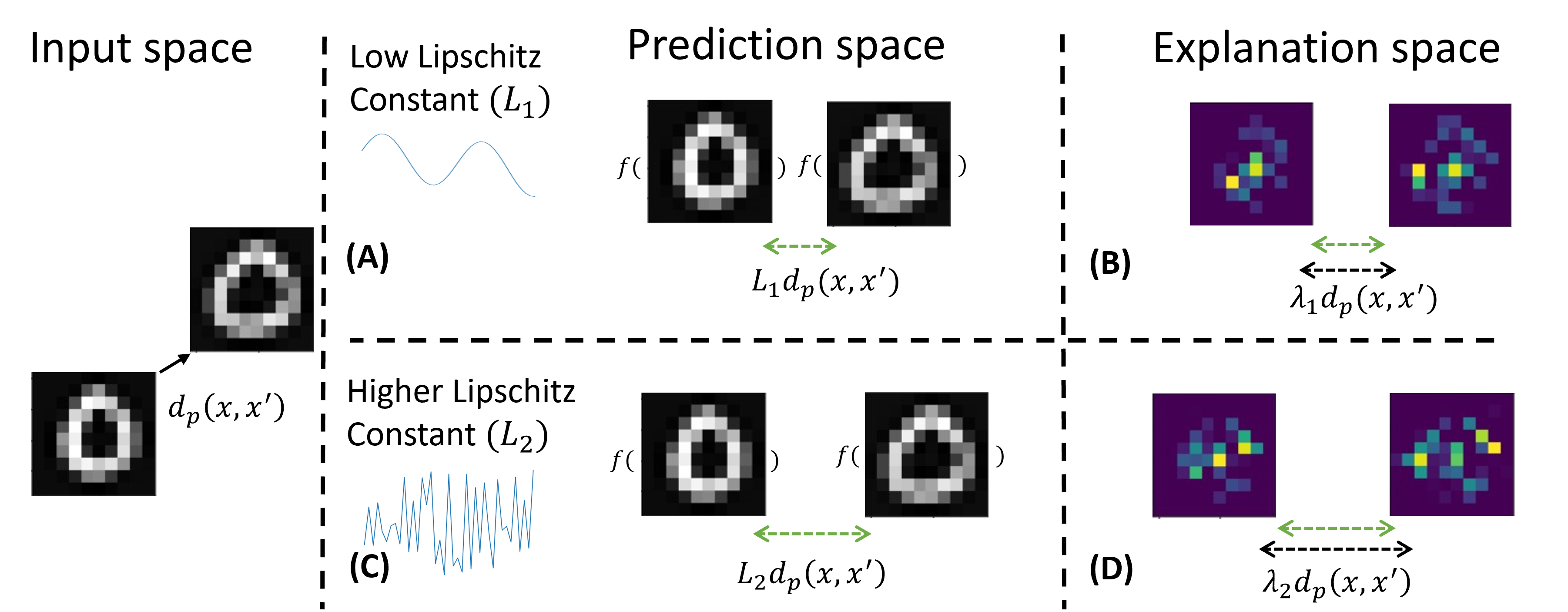}
    \caption{\textbf{Summary of our theoretical results.}  \textbf{(A)} For a black-box prediction function that is locally Lipschitz with a constant $L_1$, the predictions for any two points $x, x'$ (shown here with two similar images from MNIST dataset) such that $d_p(x,x') \leq r$ are within $L_1d_p(x,x')$ distance from each other. \textbf{(B)} Given such a prediction function, the explanation (feature attributions shown, with brighter colors indicating higher scores) for the same data points are also expected to be within $\lambda_1 d_p(x,x')$ of each other where $\lambda_1 = CL_1\sqrt{d}$ where C is a constant. \textbf{(C)} For a second black-box model with $L_2 > L_1$, our results show that \textbf{(D)} $\lambda_2 > \lambda_1$, indicating that the explanations for this black-box model can be farther apart as compared to the first prediction function. This result implies that \emph{locally smooth black-box models lend themselves to more astute explainers.}}
     \label{fig:my_label}
    \end{center}
\end{figure*}

Let $x \in \mathbb{R}^d$ be a $d$-dimensional sample from data distribution $\mathcal{D}$. We denote function $f$ as a pre-trained, black-box model. The explainer is represented by $\phi$ where $\phi(x) \in \mathbb{R}^d$ is the feature attribution vector representing attributions for all features for sample $x$; $\phi_i(x) \in \mathbb{R}$ indicates the attribution for the $i^{th}$ feature. We define $d_p(x,x')=||x - x'||_p$ as the p-norm induced distance between two points.
Our main interest is to define a metric that can capture the difference in explanations provided by an explainer to points that are close to each other in the input space. The same question has been asked for classifiers. The concept of \emph{Astuteness} was introduced by \cite{bhattacharjee2020non} in the context of classifiers; it captures the probability that similar points are assigned the same label by a classifier. Formally they provide the following definition:
    
\begin{definition}
\textit{Astuteness of classifiers} \citep{bhattacharjee2020non}: The astuteness of a classifier $f$ over $\mathcal{D}$, denoted as $A_r(f,\mathcal{D})$ is the probability that $\forall x, x' \in \mathcal{D}$ s.t. $d(x,x') \leq r$, the classifier will predict the same label.
\small\begin{equation}
    A_r(f,\mathcal{D}) = \mathbb{P}_{x,x'\sim \mathcal{D}}[f(x) = f(x') | d(x,x') \leq r]
\end{equation}
\normalsize
\end{definition}
The challenge in extending this concept of astuteness to explainers is that attribution-based explanations are not necessarily discrete; we need to generalize classifier astuteness to model continuous function outputs. With this in mind, we propose and formalize \emph{explainer astuteness}, as the probability that the explainer assigns \emph{similar} explanations to similar points.
% %\vspace{-1em}
\begin{definition} \label{def:exp_astuteness}
          \textit{Explainer astuteness}: 
        %   An Explainer $E\in \mathcal{E}$ is astute at point $x$ with radi $r$, if $E$ is robust at point $x$ with radius $r$. 
      The \emph{explainer astuteness} of an explainer $E$ over  $\mathcal{D}$, denoted as $A_{r,\lambda}(E,\mathcal{D})$ is the probability that $\forall x,x' \in \mathcal{D}$ such that $d_p(x,x') \leq r$ the explainer $E$ will provide explanations $\phi(x),\phi(x')$ that are at most $\lambda \cdot d_p(x,x')$ away from each other, where $\lambda \geq 0$
      \small\begin{equation}
      \begin{split}
           A_{r,\lambda}(E,\mathcal{D}) = \mathbb{P}_{x,x' \sim \mathcal{D}} [d_p(\phi(x), \phi(x')) \leq \lambda \cdot d_p(x, x') \\ \given[\Big] d_p(x,x') \leq r]
      \end{split}
        \label{eq:astuteness}
      \end{equation}
    \end{definition}
\normalsize
Explainer Astuteness, as defined above, captures the desirable robustness property of explainers that we are interested in. It provides a value between $0$ and $1$ that quantifies how different or similar the explanations provided by our explainer are within neighborhoods of radius $r$. $0$ implies the explainer is providing different explanations even for nearby points, and $1$ implies the explainer gives similar explanations for nearby points. 
Our goal now is to connect this measure of explainer robustness to a measure of predictor smoothness. Typically, notions of local or global \emph{detereministic} Lipschitzness are used for this purpose (see \cite{tan2023robust} and \cite{agarwal2022rethinking} for example). Instead, we choose to use \emph{probabilistic} Lipschitzness.
%as defined below:  
Probabilistic Lipschitzness captures the probability of a function being locally smooth given a radius $r$. This allows us to address predictors over the full range of local to global and non-smooth to deterministically smooth ones. This definition also mirrors our definition of explainer astuteness and helps establish the relationship between the two concepts. It is also especially useful for capturing a notion of smoothness of complicated neural network functions for which enforcing, as well as calculating, global and deterministic Lipschitzness is difficult. 
%\vspace{-1em}
\emph{Probabilistic} Lipschitzness is defined as follows:
\begin{definition}             
\label{def:plips}
\textit{Probabilistic Lipschitzness} \citep{mangal2020probabilistic}: Given $0 \leq \alpha \leq 1$, $r \geq 0$, a function $f:\mathbb{X} \rightarrow \mathbb{R}$ is probabilistically Lipschitz with a constant $L \geq 0$ if
         \small
         \small\begin{equation}
          \mathbb{P}_{x, x' \sim \mathcal{D}}[d_p(f(x),f(x')) \leq L \cdot d_p(x, x') \given[\Big] d_p(x, x') \leq r]\geq 1 - \alpha
            \label{eq:prob_lip}
         \end{equation}
             \normalsize
    \end{definition}

\section{THEORETICAL BOUNDS OF ASTUTENESS}

In this section, we theoretically show how the probabilistic Lipschitzness of black-box models relate to the astuteness of explainers. We introduce and prove theoretical bounds that connect the Lipschitz constant $L$ of the black-box model to the astuteness of three different classes of explainers: SHAP \citep{lundberg2017unified}, RISE \citep{petsiuk2018rise}, and methods that simulate individual feature removal such as CXPlain \citep{schwab2019cxplain}. An overview of these results is provided in Figure \ref{fig:my_label}.
% \vspace{-2em}
\subsection{Astuteness of SHAP}
SHAP \citep{lundberg2017unified} is one of the most popular feature attribution based explainers in use today. \citep{lundberg2017unified} unify 6 existing explanation approaches within the SHAP framework. Each of these explanation approaches (such as DeepLIFT and kernelSHAP) can be viewed as approximations of SHAP, since SHAP in its theoretical form is difficult to calculate. However, in this section we use the theoretical definition of SHAP to establish bounds on astuteness.

For notational consistency with other removal-based explainers in this work, we use an indicator vector $z \in \{0, 1\}^d$, where $z_i = 1$ when the $i^{th}$ feature is included in the subset. To indicate subsets where feature $z_i = 1$, we use $z_{+i}$; conversely, $z_{-i}$ indicates subsets where $z_i = 0$. $|z_{-i}|$ represents the number of non-zero entries in $z_{-i}$. Following definition used in \cite{lundberg2017unified} for a given data point $x \in \mathcal{X}$ and a prediction function $f$, the feature attribution provided by SHAP for the $i^{th}$ feature is given by:
    \small\begin{equation}
        \phi_i(x) = \sum_{z_{-i}} \frac{|z_{-i}|! (d - |z_{-i}| - 1)!}{d!}[f(x \odot z_{+i}) - f(x \odot z_{-i})]
    \label{eq:shap}
    \end{equation}
    \normalsize
We introduce the following Lemma (proof in Appendix~\ref{app:proof}) which is necessary for proving Theorem~\ref{thm1}.
\begin{lemma}
\label{lem1}
If,
\small\begin{equation*}
  \mathbb{P}_{x, x' \sim \mathcal{D}}[d_p(f(x),f(x')] \leq L \cdot d_p(x, x') \given[\Big] d_p(x, x') \leq r] \\ \geq 1 - \alpha
\end{equation*}
\normalsize
then for $y=x \odot z_{+i}, y'= x' \odot z_{+i}$, i.e. $y,y' \in \cup\mathbb{N}_k=\{y | y \in \mathbb{R}^d, ||y||_0 = k, y_i \neq 0\}$ for $k=1,\ldots,d$
\small\begin{equation*}
    \mathbb{P}_{x, x' \sim \mathcal{D}}[d_p(f(y),f(y')) \leq L  \cdot d_p(y,y') \given[\Big] d_p(y,y') \leq r] \\ \geq 1 - \beta
\end{equation*}
\normalsize
where $\beta \geq \alpha$ assuming that the distribution $\mathcal{D}$ is defined for all $x$ and $y$ and the equality is approached if the probability of sampling points from the set $\mathbb{N}_k=\{y | y \in \mathbb{R}^d, ||y||_0 = k, y_i \neq 0\}$ approaches zero for $k=2,\ldots,d$ relative to the probability of sampling points from $\mathbb{N}_1$ i.e. the probability of sampling points that have at least 1 element exactly equal to 0 is vanishingly small compared to the probability of sampling points that have no 0 element.
\end{lemma}
\begin{theorem}
\label{thm1}
(Astuteness of SHAP) Consider a given $r \geq 0$ and $0 \leq \alpha \leq 1$, and a trained predictive function $f$ that is probabilistic Lipschitz with a constant $L$, radius $r$ measured using $d_p(.,.)$ and with probability at least $1-\alpha$. Then for SHAP explainers we have astuteness $A_{r, \lambda} \geq 1 - \beta$ for $\lambda = 2\sqrt[p]{d}L$. Where $\beta \geq \alpha$, and $\beta \rightarrow \alpha$ under conditions specified in Lemma~\ref{lem1}.
\end{theorem}    
   %\vspace{-1em}
\begin{proof}
    Given input points $x,x'$ s.t. $d(x,x') \leq r$. And letting $\frac{|z_{-i}|! (d - |z_{-i}| - 1)!}{d!} = C_z$. Using \eqref{eq:shap} we can write,  
    \small\begin{equation}
    \begin{split}
        d_p(\phi_i(x) , & \phi_i(x')) = ||\sum_{z_{-i}}C_z[f(x \odot z_{+i}) - f(x \odot z_{-i})] \\ & -
        \sum_{z_{-i}} C_{z} [f(x' \odot z_{+i}) - f(x' \odot z_{-i})]||_p
    \end{split}
    \end{equation}
    \normalsize    
    Combining the two sums and re-arranging the R.H.S, 
    \small\begin{equation}
    \begin{split}
           d_p(\phi_i(x) , & \phi_i(x')) = ||\sum_{z_{-i}} C_z[f(x \odot z_{+i}) - f(x' \odot z_{+i}) \\ & +
           f(x' \odot z_{-i}) - f(x \odot z_{-i})]||_p
    \end{split}
    \end{equation}
    \normalsize
    Using triangular inequality on the R.H.S twice,
    \small\begin{equation}
        \begin{split}
            d_p(\phi_i(x) ,& \phi_i(x')) \leq ||\sum_{z_{-i}} C_z[f(x \odot z_{+i}) - f(x' \odot z_{+i})]||_p \\
            & + ||\sum_{z_{-i}} C_z[f(x' \odot z_{-i}) - f(x \odot z_{-i})]||_p \\
            &\leq \sum_{z_{-i}} C_z||f(x \odot z_{+i}) - f(x' \odot z_{+i})||_p \\
            & + \sum_{z_{-i}} C_z||f(x' \odot z_{-i}) - f(x \odot z_{-i})||_p
        \end{split}
    \label{eq:triangle}
    \end{equation}    \normalsize
    We can replace each value inside the sums in \eqref{eq:triangle} with the maximum value across either sums. Doing so would still preserve the inequality in \eqref{eq:triangle}, as the sum of $n$ values is always less than the maximum among those summed $n$ times.  Without loss of generality let us assume this maximum is $|f(x \odot z^*) - f(x' \odot z^*)|$ for some particular $z^*$. This gives us:
    \small\begin{equation}
    \begin{split}
       d_p(\phi_i(x) , \phi_i(x')) \leq ||f(x \odot z^*) - f(x' \odot z^*)||_p \sum_{z_{-i}} C_z \\
       + ||f(x \odot z^*) - f(x' \odot z^*)||_p \sum_{z_{-i}} C_z
    \end{split}
    \end{equation}
    \normalsize
    However, $\sum_{z_{-i}} C_z = \sum_{z_{-i}} \frac{|z_{-i}|! (d - |z_{-i}| - 1)!}{d!} = 1$, so,
    \small\begin{equation}
        \begin{split}
                    d_p(\phi_i(x) , \phi_i(x')) \leq 2||f(x \odot z^*) - f(x' \odot z^*)||_p \\
                    = 2d_p(f(x \odot z^*), f(x' \odot z^*))
        \end{split}
    \label{eq:difference}
    \end{equation}
    \normalsize
    Using the fact that $f$ is probabilistic Lipschitz with a given constant $L \geq 0$, $d_p(x,x') \leq r$, $d_p(x \odot z^*, x' \odot z^*) \leq d_p(x,x')$ and Lemma \ref{lem1}. We get:
    \small\begin{equation*}
            \mathbb{P}[2d_p(f(x \odot z^*),f(x' \odot z^*)) \leq 2L \cdot d_p(x,x')] \geq 1 - \beta
    \end{equation*}
    \normalsize
    Since \eqref{eq:difference} implies $d_p(\phi_i(x) , \phi_i(x')) \leq 2d_p(f(x \odot z^*),f(x' \odot z^*))$, the below inequality can be established:
    \small\begin{equation}
        \begin{split}
            \mathbb{P}[d_p(\phi_i(x) , \phi_i(x')) \leq 2L \cdot d_p(x,x') ] \geq 1 - \beta
        \end{split}
        \label{eq:perfeature}
    \end{equation}
    \normalsize
    Note that \eqref{eq:perfeature} is true for each feature $i \in \{1, ..., d\}$. To conclude our proof, we note that
    \small\begin{equation*}
    \begin{split}
        d_p(x,y) = \sqrt[p]{\sum_i^d|x_i - y_i|^p}\leq\sqrt[p]{\sum_i^d\max_i|x_i-y_i|^p} \\
        = \sqrt[p]{d}\max_id_p(x_i,y_i)
    \end{split}
    \end{equation*}
    \normalsize
    Utilizing this in \eqref{eq:perfeature}, without loss of generality assuming $d_p(\phi_i(x), \phi_i(x'))$ corresponds to the maximum,
    \small\begin{equation}
        \begin{split}
            \mathbb{P}[d_p(\phi(x),\phi(x')) \leq 2 \sqrt[p]{d} L \cdot d_p(x,x')] \geq 1 - \beta
        \end{split}
    \label{eq:forall}
    \end{equation}
    \normalsize
   Since $ \mathbb{P}[d_p(\phi(x),\phi(x')) \leq 2 \sqrt[p]{d} L \cdot d_p(x,x')]$ in \eqref{eq:forall} defines $A_{\lambda, r}$ for $\lambda = 2 \sqrt[p]{d}L$, this concludes the proof.
\end{proof}
\vspace{-1em}
\begin{corollary}
   If the prediction function $f$ is locally deterministically $L-$Lipschitz ($\alpha=0$) at radius $r$ then Shapley explainers are $\lambda-$astute for radius $r \geq 0$ for $\lambda = 2 \sqrt[p]{d}L$.
\end{corollary}
\vspace{-1em}
\begin{proof}
    Note that definition~\ref{def:plips} reduces to the definition of deterministic Lipschitz if $\alpha = 0$. Which means \eqref{eq:forall} will be true with probability $1$. Which concludes the proof.
\end{proof}
%\vspace{-1.5em}
\subsection{Astuteness of ``Remove Individual'' Explainers}
Within the framework of feature removal explainers, a sub-category is the explainers that work by removing a single feature from the set of all features and calculating feature attributions based on change in prediction that result from removing that feature. This category includes Occlusion, CXPlain \citep{schwab2019cxplain}, PredDiff \citep{zintgraf2017visualizing} Permutation tests \citep{strobl2008conditional}, and feature ablation explainers \citep{lei2018distribution}.

``Remove individual'' explainers determine feature explanations for the $i^{th}$ feature by calculating the difference in prediction with and without that feature included  for a given point $x$. Let $z_{-i}\in\{0,1\}^d$ represent a binary vector with $z_i=0$, then the explanation for feature $i$ can be written as:
\small\begin{equation}
   \phi(x_i) = f(x) - f(x \odot z_{-i})
\label{eq:removal_exp}
\end{equation}
\normalsize
\begin{theorem}
\label{thm3}
(Astuteness of Remove individual explainers) Consider a given $r\geq0$ and $0 \leq \alpha \leq 1$ and a trained predictive function $f$ that is locally probabilistic Lipschitz with a constant $L$, radius $r$ measured using $d_p(.,.)$ and probability at least $1-\alpha$. Then for Remove individual explainers, we have the astuteness $A_{r,\lambda} \geq 1 - \alpha$, for $\lambda = 2\sqrt[p]{d}L$, where d is the dimensionality of the data.
\end{theorem}
%%\vspace{-1em}
\begin{proof}
(Sketch, full proof in Appendix~\ref{app:proof}) By considering another point $x'$ such that $d_p(x,x') \leq r$ and \eqref{eq:removal_exp} we get,
\normalsize
\small\begin{equation}
    d_p(\phi(x_i), \phi(x'_i)) = d_p(f(x) - f(x \odot z_{-i}), f(x') - f(x' \odot z_{-i}))
\end{equation}
\normalsize

   then following the exact same steps as the proof for Theorem \ref{thm1}, i.e. writing the right hand side in terms of $p$-norm and utilizing triangular inequality, leads us to the desired result.
   \end{proof}
  \begin{corollary}
  If the prediction function $f$ is locally $L-$Lipschitz at radius $r \geq 0$, then remove individual explanations are $\lambda-$astute for radius $r$ and $\lambda = 2\sqrt[p]{d}L$.
  \end{corollary} %%%\vspace{-1em}
  \begin{proof}
  Same as proof for Corollary 2.1.
  \end{proof}

%%%%%%%%%%%%%%%%%%%%%%%%%%%%%%%%%%%%%%%%%%%%%%%%%%%%%%%%
\subsection{Astuteness of RISE}
 RISE determines feature explanation for the $i^{th}$ feature by sampling subsets of features and then calculating the mean value of the prediction function when feature $i$ is included in the subset. RISE feature attribution for a given point $x$ and feature $i$ for a prediction function $f$ can be written as:
   \normalsize
   \small\begin{equation}
       \phi_i(x) = \mathbb{E}_{p(z|z_i=1)}[f(x \odot z)]
    \label{eq:rise}
   \end{equation}
    \normalsize
The following theorem establishes the bound on $\lambda$ for \emph{explainer astuteness} of RISE in relation to the Lipschitzness of black-box prediction function.
\begin{theorem}
\label{thm2}
(Astuteness of RISE) Consider a given $r \geq 0$ and $0 \leq \alpha \leq 1$, and a trained predictive function $f$ that is locally deterministically Lipschitz with a constant $L$ ($\alpha=0$), radius $r$ measured using $d_p(.,.)$ and probability at least $1-\alpha$. Then for RISE explainer is $\lambda-$astute for radius $r$ and $\lambda = \sqrt[p]{d}L$.
\end{theorem}

   %\vspace{-1em}

\begin{proof}(Sketch, full proof in Appendix~\ref{app:proof})

    Given inputs $x, x'$ s.t. $d(x,x') \leq r$, using Eq. \eqref{eq:rise},
   \normalsize
   \small\begin{equation}
    \begin{split}
    &d_p(\phi_i(x),\phi_i(x')) \\ & =d_p(\mathbb{E}_{p(z|z_i=1)}[f(x \odot z)],\mathbb{E}_{p(z|z_i=1)}[f(x' \odot z)]) \\
    & =||\mathbb{E}_{p(z|z_i=1)}[f(x \odot z)]-\mathbb{E}_{p(z|z_i=1)}[f(x' \odot z)]||_p \\
    & =||\mathbb{E}_{p(z|z_i=1)}[f(x \odot z) - f(x' \odot z)]||_p 
    \end{split}
   \end{equation}
   \normalsize
   Using Jensen's inequality on R.H.S and $E[f] \leq \max f$
    \small\begin{equation}
            d_p(\phi_i(x),\phi_i(x')) \leq \max_z d_p(f(x \odot z),f(x' \odot z))
       \label{eq:risep1}
   \end{equation}
   \normalsize
   $f$ is is deterministically Lipschitz and $d_p(\phi(x), \phi(x')) \leq \sqrt[p]{d} * \max_i d_p(\phi_i(x),\phi_i(x'))$, this gives us,
   \normalsize
   \small\begin{equation}
       \mathbb{P}[ d_p(\phi(x),\phi(x') \leq \sqrt[p]{d}L \cdot d_p(x,x')] \geq 1
       \label{eq:riseall}
   \end{equation}
   \normalsize
   Since $\mathbb{P}[d_p(\phi(x),\phi(x') \leq \sqrt[p]{d}L \cdot d_p(x,x')]$ defines $A_{\lambda,r}$ for $\lambda = \sqrt[p]{d}L$, this concludes the proof.
\end{proof}
\subsection{Connecting Explainer Astuteness and Probabilistic Lipschitzness} \label{subsec:impl}
% \vspace{-2mm}
The above theoretical results for the three classes of explainers provide the same critical implication, that is, explainer astuteness is lower bounded by the probabilistic Lipschitzness of the prediction function. This means that black-box classifiers that are locally smooth (have a small $L$ at a given radius $r$) lend themselves to probabilistically more robust explanations. \emph{This work provides the theoretical support on the importance of enforcing smoothness of classifiers to astuteness of explanations}. Note that while this implication makes intuitive sense, proving it for specific explainers is non-trivial as demonstrated by the three theorems above and their respective proofs. The statement holds true for all three explainers when the classifier can be assumed to be deterministically Lipschitz, the conditions under which it is still true for probabilistic Lipschitzness vary in each case. For Theorem~\ref{thm1} we have to assume that distribution $\mathcal{D}$ is defined over masked data in addition to the input data and ideally the probability of sampling of masked data from is significantly smaller compared to probability of sampling points with no value exactly equal to 0. For Theorem~\ref{thm3} the statement is true without additional assumptions. For Theorem~\ref{thm2} we can only prove the statement to be true for the deterministic case.
% \vspace{-2mm}
\section{EXPERIMENTS} \label{sec:experiments}
% \vspace{-2mm}

\begin{figure*}[t]
  % \begin{minipage}[c]{0.75\textwidth}
    \centering    \includegraphics[width=0.85\textwidth]{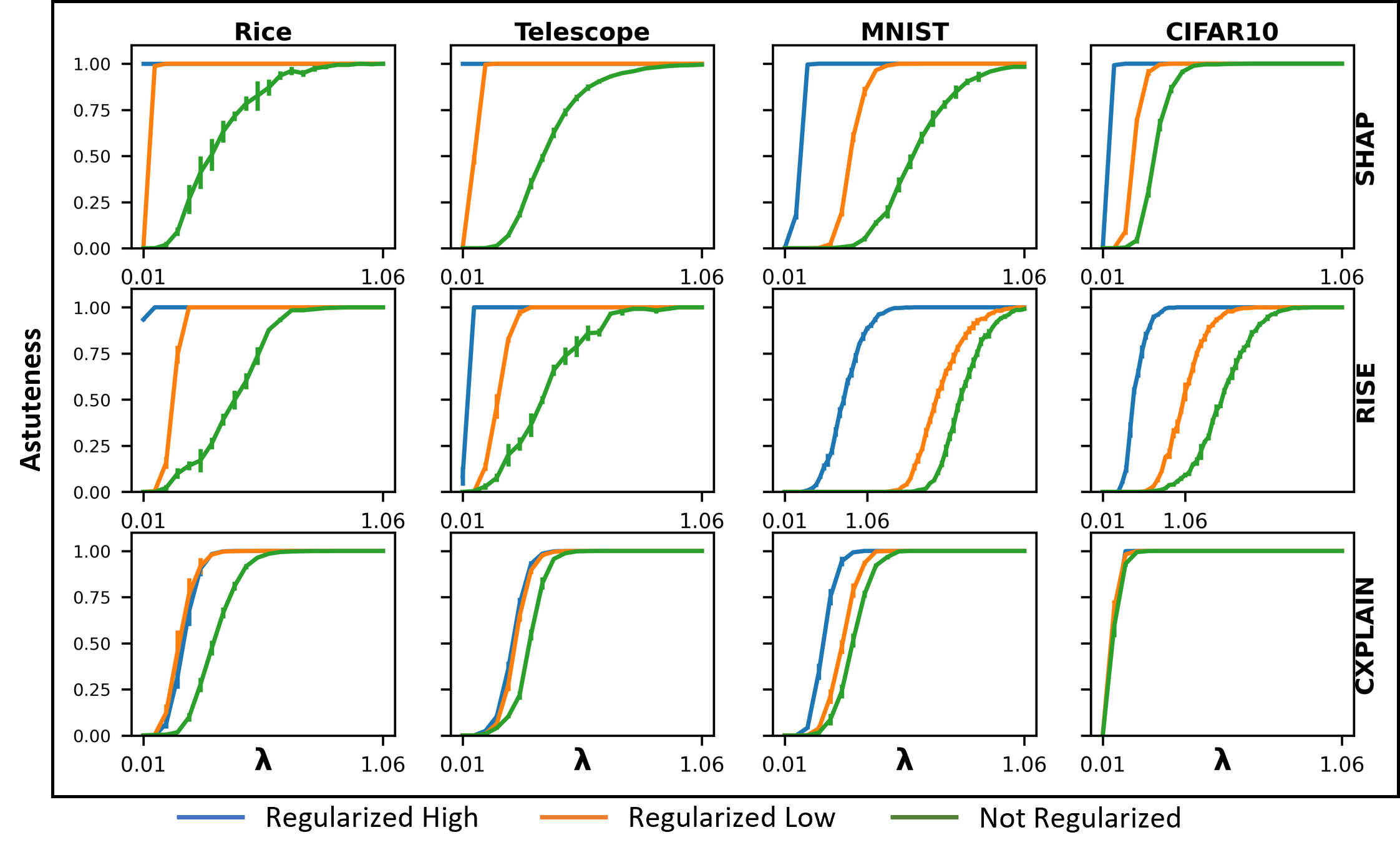}
    \setlength{\abovecaptionskip}{-3pt}
    \caption{\textbf{Smooth functions result in astute explanations. }Regularizing the Lipschitzness of a neural network during training results in higher astuteness for the same value of $\lambda$. Higher regularization results in lower Lipschitz constant \cite{gouk2021regularisation}. Astuteness reaches $1$ for smaller values of $\lambda$ with Lipschitz regularized training, as expected from our theorems. The errorbars represent results across 5 runs to account for randomness in explainer runs.} \label{fig:experiments2}
  % \end{minipage}
\end{figure*}
To demonstrate the validity of our theoretical results, we perform a series of experiments. We train four different classifiers on each of five datasets, and then explain the decisions of these classifiers using three explainers and calculate astuteness of these explainers. We use $p=2$ in all experiments.

\textbf{Datasets.} We utilize three simulated datasets introduced by \cite{chen2018learning} namely \emph{Orange Skin}(OS), \emph{Nonlinear Additive}(NA) and \emph{Switch}, and two real world datasets from UCI Machine Learning repository \citep{asuncion2007uci} namely \emph{Rice} \citep{cinar2019classification} and \emph{Telescope} \citep{ferenc2005magic} as well as \emph{MNIST}\citep{mnist} and \emph{CIFAR10}\citep{cifar10} datasets. See Appendix~\ref{app:datasets} for details.

\textbf{Classifiers.} For each dataset we train the following four classifiers; a $2$-\emph{layer} MLP, a $4$-\emph{layer} MLP, a \emph{linear} classifier, and a \emph{svm}. For training Details see Appendix~\ref{app:training_details}. 
% We perform two sets of experiments, first is to show the effect of Lipschitz constraints on explainer astuteness, and second to show that the probabilistic Lipschitzness of classifiers can indeed be used to lower bound explainer astuteness as predicted by our theorems.

\textbf{Explainers.} We evaluate 3 explainers, one for each of our theorems. Gradient-based approximation and kernel shap approximation of \textbf{SHAP}\citep{lundberg2017unified} for the NN classifiers and SVM, respectively, serve as representative of Theorem~\ref{thm1}. We modify the implementation of \textbf{RISE} \citep{petsiuk2018rise} provided by the authors to work with tabular datasets; this serves as representative for Theorem~\ref{thm2}. The implementation for \textbf{CXPlain} \citep{schwab2019cxplain} serves as representative for Theorem~\ref{thm3}.\footnote{SHAP: \url{https://github.com/slundberg/shap}, RISE: \url{https://github.com/eclique/RISE}, CXPLAIN: \url{https://github.com/d909b/cxplain}}.

\begin{table*}[t]
    {\centering
    % This represents the average tightness of the lower bound.
    \setlength{\belowcaptionskip}{-20pt}
    \scriptsize
    % \small
    \begin{center}
    \setlength\tabcolsep{4pt}
        \begin{tabular}{lccccccccccccc}
        \toprule
        & \multicolumn{3}{c}{2layer}
        & \multicolumn{3}{c}{4layer}
        & \multicolumn{3}{c}{linear}
        & \multicolumn{3}{c}{svm}\\
        \cmidrule(lr){2-4}
        \cmidrule(lr){5-7}
        \cmidrule(lr){8-10}
        \cmidrule(lr){11-13}
            Datasets      
            & SHAP
            & RISE
            & CXPlain
            & SHAP
            & RISE
            & CXPlain      
            & SHAP
            & RISE
            & CXPlain      
            & SHAP
            & RISE
            & CXPlain
            \\           
        \midrule
            	OS
            	& .585
            	& .477
            	& .551
            	& .489
            	& .415
            	& .426
            	& .043
            	& .017
            	& .043
            	& .761
            	& .628
            	& .732
            	\\   
            	NA
            	& .359
            	& .289
            	& .318
            	& .285
            	& .216
            	& .244
            	& .452
            	& .391
            	& .474
            	& .742
            	& .653
            	& .708
            	\\             	
            	Switch
            	& .053
            	& .053
            	& .003
            	& .086
            	& .083
            	& .039
            	& .043
            	& .028
            	& .034
            	& .557
            	& .472
            	& .524
            	\\            	
            	Rice
            	& .249
            	& .142
            	& .229
            	& .292
            	& .131
            	& .252
            	& .258
            	& .165
            	& .241
            	& .426
            	& .347
            	& .413
            	\\             	
            	Telescope
            	& .324
            	& .213
            	& .317
            	& .345
            	& .244
            	& .333
            	& .223
            	& .149
            	& .211
            	& .501
            	& .439
            	& .504
            	\\             	
            	CIFAR10
            	& .333
            	& .238
            	& .337
            	& .350
            	& .224
            	& .356
            	& .340
            	& .235
            	& .345
            	& .336
            	& .329
            	& .255           \\  	
            	MNIST
            	& .441
            	& .297
            	& .435
            	& .522
            	& .357
            	& .519
            	& .455
            	& .303
            	& .448
            	& .443
            	& .379
            	& .448
            	\\     
        \bottomrule  
        \end{tabular}
    \end{center}
    }
    \caption{$\mathbf{AUC_{emp} - AUC_{pred}}(\downarrow)$. The observed AUC ($\textrm{AUC}_{\textrm{emp}}$) is lower bounded by the predicted AUC ($\textrm{AUC}_{\textrm{pred}}$). As expected, the difference between the two is always $\geq 0$.} 
    \label{table:auc}

    \end{table*}

%%%%%%%%%%%%%%%%%%%%%%%%%%%%%%%%%%%%%%%%%%%%%%%%%%%%%%%%%%%%%%%%%%%%%%
% \vspace{-2mm}
\subsection{Effect of Lipschitz Constraints on Explainer Astuteness}
% \vspace{-2mm}
In this experiment, we utilize Lipschitz-constrained classifiers introduced by \cite{gouk2021regularisation}, which constrain the Lipschitz constant for each layer by adding a projection step during training.
During each gradient update, the weight matrices undergo a projection onto a feasible set if they violate the constraints imposed on the Lipschitz constant. The level of constraint can be adjusted through a hyperparameter, allowing control over the impact on the weight matrices.  We use this method to train a four-layer MLP with high, low, and no Lipschitz constraint. We then calculate astuteness of each of our explainers for all three versions of this neural network. Figure~\ref{fig:experiments2} shows the results. The goal of this experiment is to demonstrate the relationship between the Lipschitzness of a NN and the astuteness of explainers. As the \textit{same} NN is trained on the \textit{same} data but with different levels of Lipschitz constraints enforced, the astuteness of explainers varies accordingly. In all cases we see astuteness reaching $1$ for smaller values of $\lambda$ for the same NN when it is highly constrained (lower Lipschitz constant $L$) vs less constrained or unconstrained. \emph{The results provide empirical evidence in support of the main conclusion of our work: i.e., enforcing Lipschitzness on classifiers lends them to more astute post-hoc explanations.}
%%%%%%%%%%%%%%%%%%%%%%%%%%%%%%%%%%%%%%%%%%%%%%%%%%%%%%%%%%%%%%%%%%%%%%
\vspace{-2mm}
\subsection{Estimating Probabilistic Lipschitzness and Lower Bound for Astuteness} \label{sec:estimatingL}
\vspace{-2mm}
To demonstrate the connection between explainer astuteness and probabilistic Lipschitzness as shown in our theorems, we estimate the probabilistic Lipschitzness of various classifiers. We achieve this by empirically estimating $\mathbb{P}_{x,x' \sim \mathcal{D}}$ \eqref{eq:prob_lip} for a range of values of $L \in (0, 1)$ incremented by $0.1$. This is done for each classifier and for each dataset $D$, and we set $r$ as the median of pairwise distance for all training points.
According to \eqref{eq:prob_lip}, this gives us an upper bound on $1-\alpha$, i.e. we can say that for a given $L, r$ the classifier is Lipschitz with probability at least $1 - \alpha$.
We can use the estimates for probabilistic Lipschitzness to predict the lower bound of astuteness using our theorems.
Theorems \ref{thm1}, \ref{thm3}, \ref{thm2} imply that for $\lambda = CL\sqrt{d}$, explainer astuteness is $\geq 1 - \alpha$. This indicates that for $\lambda \geq LC\sqrt{d}$, explainer astuteness should be lower-bounded by $1 - \alpha$. For each dataset-classifier-explainer combination we can plot two curves.
First, the empirical estimation of explainer astuteness using Definition~\ref{def:exp_astuteness}.
Second, the curve that represents the predicted lower bound on explainer astuteness given a classifier, as just described.
According to our theoretical results, at a given $\lambda$, the estimated explainer astuteness should stay above the predicted astuteness based on the Lipschitzness of classifiers.
Table \ref{table:auc} shows the difference between the area under the curve (AUC) for the empirically calculated astuteness ($\mathbf{AUC_{emp}}$) and the predicted lower bound ($\mathbf{AUC_{pred}}$).
This number captures the average difference of the lower bound over a range of $\lambda$ values. \emph{Note that the values are all positive supporting our result as a lower bound.}
The associated curves are shown in App. \ref{app:results} Figure \ref{fig:sup_experiments}.
% \vspace{-2mm}
\subsection{Astuteness as a Metric} \label{sec:astuteness_experiment}
\begin{figure}[t]
    \centering   \includegraphics[width=\columnwidth]{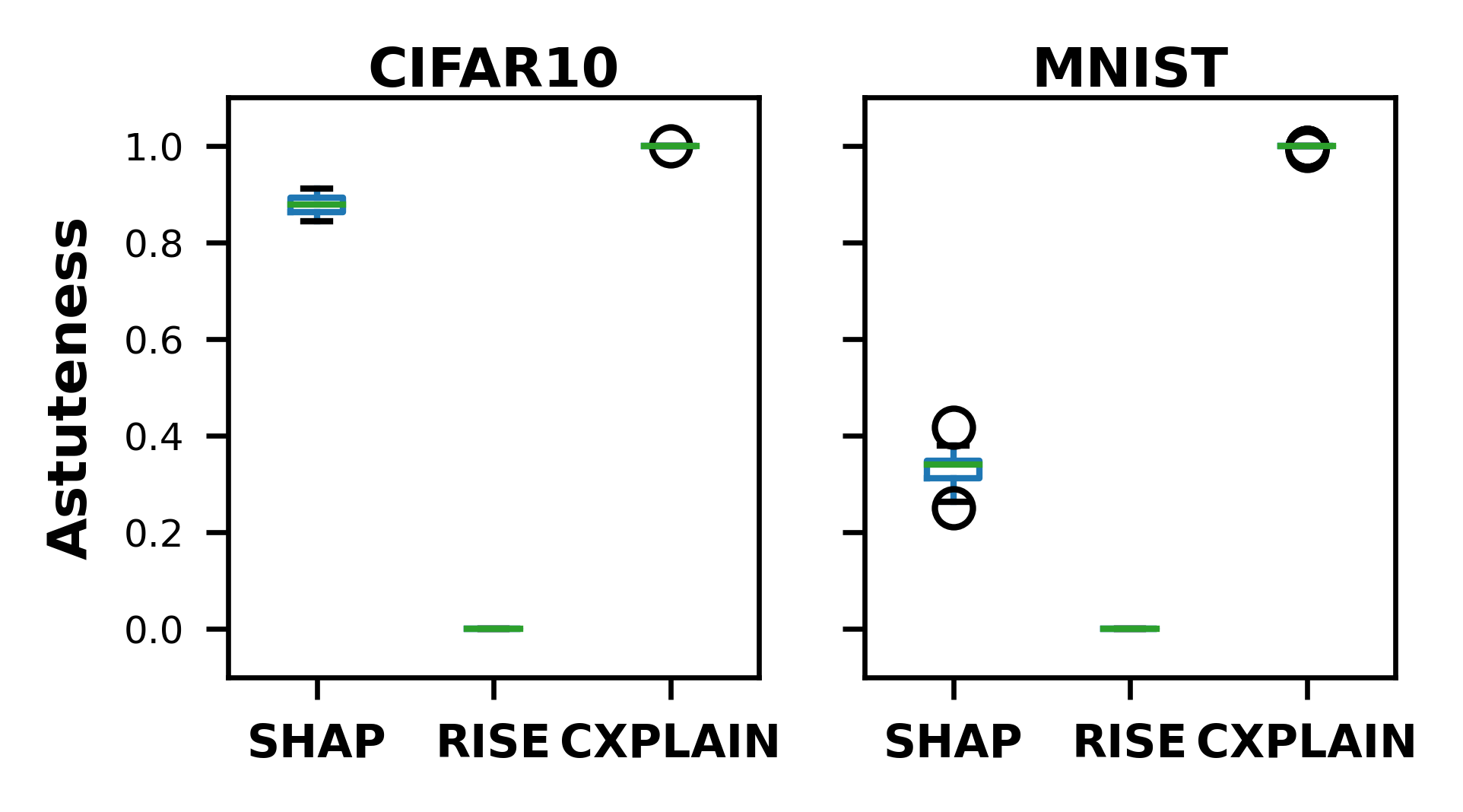}
    \setlength{\abovecaptionskip}{-15pt}
    \caption{\textbf{Astuteness as a metric.} Different explainers display different levels of astuteness. In our experiments, RISE consistently displayed lower astuteness for the same value of $\lambda$ compared to SHAP and CXPLAIN. This indicates that among these three , on the considered datasets and classifiers, RISE is least robust and CXPlain is the most robust.
    Results are shown across 20 explainer runs.} \label{fig:experiments3}
\end{figure}

% \vspace{-2mm}
Our proposed metric of \emph{Explainer Astuteness} (Definition \ref{def:exp_astuteness}) enables the comparison of different explainers based on the expected robustness of their explanations. This allows users to select a more astute explainer in situations where robust explanations are required. In Figure \ref{fig:experiments3} we compare the astuteness of SHAP, RISE, and CXPlain. We observe that RISE is consistently less astute on the evaluated datasets, in contrast to CXPlain, which exhibits high astuteness. Therefore in this case CXPlain would be preferable for obtaining robust explanations. 

We also provide results on four datasets from OpenXAI benchmark \citep{agarwal2022openxai}, a synthetic data from OpenXAI (O-Synthetic), Home Equity Line of Credit (HELOC) \citep{holter2018fico}, Propublica's COMPAS dataset \citep{jordan2015effect}, and Adult Income (Adult) dataset \citep{yeh2009comparisons}. Reported in Table~\ref{table:openxai}, we calculate astuteness over a range of values of $\lambda \in \{0.1, 1.1\}$ in increments of $0.1$, and report the AUC of astuteness over this range for SHAP and LIME using the implementations and pre-trained models made available on OpenXAI's github \footnote{\url{https://github.com/AI4LIFE-GROUP/OpenXAI}}, higher AUC indicates higher astuteness over the range. For each dataset-explainer combination we also report a metric for measuring stability - Relative Input Stability (RIS) - a measure of maximum change in explanation relative to changes in input \citep{alvarez2018robustness} and a measure of faithfulness -  Prediction Gap on
Important feature perturbation (PGI),  which computes the difference in prediction probability that results from perturbing the features deemed as influential by a given post hoc explanation \citep{dai2022fairness}. These metrics and their values are reported on the OpenXAI leaderboard \footnote{\url{https://open-xai.github.io/leaderboard}}. Since astuteness is closely related to stability, we can expect the astuteness AUC to be higher for the explainer that has a higher RIS for each dataset-predictor combination, as evident from bolded values in \ref{table:openxai} that is indeed generally the case. 
% Astuteness improves the ability for users to evaluate explainers that other metrics (e.g. infidelity \citep{yeh2019fidelity} or sensitivity \citep{yin2021faithfulness}) are unable to capture.

% \vspace{-1em}
\begin{table*}[t]
    {\centering
    % This represents the average tightness of the lower bound.
    \setlength{\belowcaptionskip}{-20pt}
    \scriptsize
    % \small
    \begin{center}
    \setlength\tabcolsep{4pt}
        \begin{tabular}{lcccccc|cccccc}
        \toprule
        & \multicolumn{6}{c|}{Neural Network}
        & \multicolumn{6}{c}{Logistic Regression}\\
        \cmidrule(lr){2-7}
        \cmidrule(lr){8-13}
            & \multicolumn{3}{c}{SHAP}
            & \multicolumn{3}{c|}{LIME}
            & \multicolumn{3}{c}{SHAP}
            & \multicolumn{3}{c}{LIME}\\
        \cmidrule(lr){2-4}
        \cmidrule(lr){5-7}
        \cmidrule(lr){8-10}
        \cmidrule(lr){11-13}
            Datasets      
            & AUC $\uparrow$
            & PGI $\uparrow$     
            & RIS $\downarrow$     
            & AUC $\uparrow$
            & PGI $\uparrow$     
            & RIS $\downarrow$      
            & AUC $\uparrow$
            & PGI $\uparrow$     
            & RIS $\downarrow$      
            & AUC $\uparrow$
            & PGI $\uparrow$     
            & RIS $\downarrow$
            \\           
        \midrule
            	O-Synthetic
            	& .198 $\pm$ .001
            	& \textbf{.320}
            	& 5.96
            	& \textbf{.379 $\pm$ .001}
            	& .250
            	& \textbf{5.53}
            	& \textbf{.209 $\pm$ .000}
            	& \textbf{.171}
            	& \textbf{5.67}
            	& .029 $\pm$ .000
            	& .154
            	& 9.35
            	\\   
            	HELOC
            	& .130 $\pm$ .000
            	& \textbf{.260}
            	& 1.57
            	& \textbf{.506 $\pm$ .002}
            	& \textbf{.260}
            	& \textbf{1.19}
            	& \textbf{.177 $\pm$ .000}
            	& .121
            	& \textbf{1.46}
            	& .045 $\pm$ .002
            	& \textbf{.156}
            	& 4.53
            	\\             	
            	COMPAS
            	& .276 $\pm$ .001
            	& \textbf{.274}
            	& \textbf{4.24}
            	& \textbf{.515 $\pm$ .000}
            	& .232
            	& 4.49
            	& .160 $\pm$ .000
            	& \textbf{.128}
            	& \textbf{3.12}
            	& \textbf{.352 $\pm$ .001}
            	& .108
            	& 4.24
            	\\             	
            	Adult
            	& .085 $\pm$ .000
            	& .53
            	& 1.98
            	& \textbf{.206 $\pm$ .000}
            	& \textbf{.670}
            	& \textbf{1.79}
            	& \textbf{.114 $\pm$ .000}
            	& .391
            	& 1.86
            	& .042 $\pm$ .001
            	& \textbf{.420}
            	& \textbf{1.72}
             \\
        \bottomrule  
        \end{tabular}
    \end{center}
    }
    \caption{\textbf{Metrics for post-hoc explanations.} The area under the curve (AUC)$\uparrow$ of astuteness for SHAP and LIME on benchmark datasets from OpenXAI. Prediction Gap on Important feature perturbation (PGI)$\uparrow$ is a measure of faithfulness, and Relative Input Stablity (RIS)$\downarrow$ is a measure of stability. Both PGI and RIS metrics are reported from OpenXAI leaderboard. } 
    \label{table:openxai}

    \end{table*}

% \vspace{-2mm}
\section{CONCLUSION AND LIMITATIONS} \label{sec:conclusion}
% \vspace{-3mm}
In this paper we formalize \emph{explainer astuteness}, which captures the ability of an explainer to assign similar explanations to similar points. We prove that this explainer astuteness is proportional to the \emph{probabilistic Lipschitzness} of the predictor that is being explained. As probabilistic Lipschitzness captures the local smoothness of a function, this result suggests that enforcing smoothness on predictors can lend them to more robust explanations.

Regarding limitations, we observe that our empirical results suggest that our predicted lower bound can be tightened further. One possible explanation is that the tightness of this bound depends on how different explainers calculate attribution scores, e.g. empirically we observe RISE and SHAP (that both depend on expectations over subsets) behave similarly to each other but different from CXPlain.
Some explainers, such as LIME for tabular data, have an optional discretization step when calculating feature attributions. As a consequence, two observations with all features belonging to the same bins would receive exactly the same explanation, whereas two arbitrarily close inputs may receive completely different explanations (when the number of perturbed sample is large \citep{garreau2020looking}). In that sense, tabular LIME would not be astute by our formulation, regardless of classifier Lipschitzness.
% Robustness is also only one property of a reliable explainer; other properties have been investigated in recent literature, as we outline in Section \ref{sec:intro}. These other properties, e.g. faithfulness \citep{agarwal2022probing} may also be theoretically probed akin to our analysis for robustness.
Additionally, the usefulness of explainer robustness or astuteness is application-dependant. For example, robustness can sometimes be at odds with correctness (See for example \citep{zhou2022exsum} and ``Logic Trap 3'' in \citep{ju2022logic}) and is best viewed as one part of explanation reliability and trustworthiness \citep{zhou2022exsum}.

From a broader societal impact perspective, we would like to make it clear that just enforcing Lipschitzness on black-box classifiers is not enough in terms of making them more transparent and interpretable. Our work is intended to be a call to action for the field to concentrate more on improving black-box models for explainability purposes from the very start and provides one of many ways to achieve that goal.

\subsubsection*{Acknowledgements}
This work was supported in part by grants NIH 2T32HL007427-41 and U01HL089856 from the National Heart, Lung, and Blood Institute and by Northeastern University's Institute for Experiential AI.

% REFERENCES

% \clearpage
\bibliography{main}

\begin{thebibliography}{58}
\providecommand{\natexlab}[1]{#1}
\providecommand{\url}[1]{\texttt{#1}}
\expandafter\ifx\csname urlstyle\endcsname\relax
  \providecommand{\doi}[1]{doi: #1}\else
  \providecommand{\doi}{doi: \begingroup \urlstyle{rm}\Url}\fi

\bibitem[Agarwal et~al.(2022{\natexlab{a}})Agarwal, Johnson, Pawelczyk, Krishna, Saxena, Zitnik, and Lakkaraju]{agarwal2022rethinking}
Chirag Agarwal, Nari Johnson, Martin Pawelczyk, Satyapriya Krishna, Eshika Saxena, Marinka Zitnik, and Himabindu Lakkaraju.
\newblock Rethinking stability for attribution-based explanations.
\newblock \emph{arXiv preprint arXiv:2203.06877}, 2022{\natexlab{a}}.

\bibitem[Agarwal et~al.(2022{\natexlab{b}})Agarwal, Krishna, Saxena, Pawelczyk, Johnson, Puri, Zitnik, and Lakkaraju]{agarwal2022openxai}
Chirag Agarwal, Satyapriya Krishna, Eshika Saxena, Martin Pawelczyk, Nari Johnson, Isha Puri, Marinka Zitnik, and Himabindu Lakkaraju.
\newblock Openxai: Towards a transparent evaluation of model explanations.
\newblock \emph{Advances in Neural Information Processing Systems}, 35:\penalty0 15784--15799, 2022{\natexlab{b}}.

\bibitem[Agarwal et~al.(2022{\natexlab{c}})Agarwal, Zitnik, and Lakkaraju]{agarwal2022probing}
Chirag Agarwal, Marinka Zitnik, and Himabindu Lakkaraju.
\newblock Probing gnn explainers: A rigorous theoretical and empirical analysis of gnn explanation methods.
\newblock In \emph{International Conference on Artificial Intelligence and Statistics}, pages 8969--8996. PMLR, 2022{\natexlab{c}}.

\bibitem[Agarwal et~al.(2021)Agarwal, Jabbari, Agarwal, Upadhyay, Wu, and Lakkaraju]{agarwal2021towards}
Sushant Agarwal, Shahin Jabbari, Chirag Agarwal, Sohini Upadhyay, Zhiwei~Steven Wu, and Himabindu Lakkaraju.
\newblock Towards the unification and robustness of perturbation and gradient based explanations.
\newblock \emph{arXiv preprint arXiv:2102.10618}, 2021.

\bibitem[Alemi et~al.(2016)Alemi, Fischer, Dillon, and Murphy]{alemi2016deep}
Alexander~A Alemi, Ian Fischer, Joshua~V Dillon, and Kevin Murphy.
\newblock Deep variational information bottleneck.
\newblock \emph{arXiv preprint arXiv:1612.00410}, 2016.

\bibitem[Alvarez-Melis and Jaakkola(2018)]{alvarez2018robustness}
David Alvarez-Melis and Tommi~S Jaakkola.
\newblock On the robustness of interpretability methods.
\newblock \emph{arXiv preprint arXiv:1806.08049}, 2018.

\bibitem[Arrieta et~al.(2020)Arrieta, D{\'\i}az-Rodr{\'\i}guez, Del~Ser, Bennetot, Tabik, Barbado, Garc{\'\i}a, Gil-L{\'o}pez, Molina, Benjamins, et~al.]{arrieta2020explainable}
Alejandro~Barredo Arrieta, Natalia D{\'\i}az-Rodr{\'\i}guez, Javier Del~Ser, Adrien Bennetot, Siham Tabik, Alberto Barbado, Salvador Garc{\'\i}a, Sergio Gil-L{\'o}pez, Daniel Molina, Richard Benjamins, et~al.
\newblock Explainable artificial intelligence (xai): Concepts, taxonomies, opportunities and challenges toward responsible ai.
\newblock \emph{Information Fusion}, 58:\penalty0 82--115, 2020.

\bibitem[Asuncion and Newman(2007)]{asuncion2007uci}
Arthur Asuncion and David Newman.
\newblock Uci machine learning repository, 2007.

\bibitem[Aziznejad et~al.(2020)Aziznejad, Gupta, Campos, and Unser]{aziznejad2020deep}
Shayan Aziznejad, Harshit Gupta, Joaquim Campos, and Michael Unser.
\newblock Deep neural networks with trainable activations and controlled lipschitz constant.
\newblock \emph{IEEE Transactions on Signal Processing}, 68:\penalty0 4688--4699, 2020.

\bibitem[Bhattacharjee and Chaudhuri(2020)]{bhattacharjee2020non}
Robi Bhattacharjee and Kamalika Chaudhuri.
\newblock When are non-parametric methods robust?
\newblock In \emph{International Conference on Machine Learning}, pages 832--841. PMLR, 2020.

\bibitem[Chen et~al.(2018)Chen, Song, Wainwright, and Jordan]{chen2018learning}
Jianbo Chen, Le~Song, Martin Wainwright, and Michael Jordan.
\newblock Learning to explain: An information-theoretic perspective on model interpretation.
\newblock In \emph{International Conference on Machine Learning}, pages 883--892. PMLR, 2018.

\bibitem[Cinar and Koklu(2019)]{cinar2019classification}
Ilkay Cinar and Murat Koklu.
\newblock Classification of rice varieties using artificial intelligence methods.
\newblock \emph{International Journal of Intelligent Systems and Applications in Engineering}, 7\penalty0 (3):\penalty0 188--194, 2019.

\bibitem[Covert et~al.(2020)Covert, Lundberg, and Lee]{covert2020explaining}
Ian Covert, Scott Lundberg, and Su-In Lee.
\newblock Explaining by removing: A unified framework for model explanation.
\newblock \emph{arXiv preprint arXiv:2011.14878}, 2020.

\bibitem[Dai et~al.(2022)Dai, Upadhyay, Aivodji, Bach, and Lakkaraju]{dai2022fairness}
Jessica Dai, Sohini Upadhyay, Ulrich Aivodji, Stephen~H Bach, and Himabindu Lakkaraju.
\newblock Fairness via explanation quality: Evaluating disparities in the quality of post hoc explanations.
\newblock In \emph{Proceedings of the 2022 AAAI/ACM Conference on AI, Ethics, and Society}, pages 203--214, 2022.

\bibitem[Dombrowski et~al.(2019)Dombrowski, Alber, Anders, Ackermann, M\"{u}ller, and Kessel]{dombrowski_explanations_can_be_manipulated}
Ann-Kathrin Dombrowski, Maximillian Alber, Christopher Anders, Marcel Ackermann, Klaus-Robert M\"{u}ller, and Pan Kessel.
\newblock Explanations can be manipulated and geometry is to blame.
\newblock In H.~Wallach, H.~Larochelle, A.~Beygelzimer, F.~d\textquotesingle Alch\'{e}-Buc, E.~Fox, and R.~Garnett, editors, \emph{Advances in Neural Information Processing Systems}, volume~32. Curran Associates, Inc., 2019.
\newblock URL \url{https://proceedings.neurips.cc/paper_files/paper/2019/file/bb836c01cdc9120a9c984c525e4b1a4a-Paper.pdf}.

\bibitem[Fawzi et~al.(2017)Fawzi, Moosavi-Dezfooli, and Frossard]{fawzi2017robustness}
Alhussein Fawzi, Seyed-Mohsen Moosavi-Dezfooli, and Pascal Frossard.
\newblock The robustness of deep networks: A geometrical perspective.
\newblock \emph{IEEE Signal Processing Magazine}, 34\penalty0 (6):\penalty0 50--62, 2017.

\bibitem[Fazlyab et~al.(2019)Fazlyab, Robey, Hassani, Morari, and Pappas]{fazlyab2019efficient}
Mahyar Fazlyab, Alexander Robey, Hamed Hassani, Manfred Morari, and George Pappas.
\newblock Efficient and accurate estimation of lipschitz constants for deep neural networks.
\newblock \emph{Advances in Neural Information Processing Systems}, 32, 2019.

\bibitem[Fel et~al.(2022)Fel, Vigouroux, Cad{\`e}ne, and Serre]{fel2022good}
Thomas Fel, David Vigouroux, R{\'e}mi Cad{\`e}ne, and Thomas Serre.
\newblock How good is your explanation? algorithmic stability measures to assess the quality of explanations for deep neural networks.
\newblock In \emph{Proceedings of the IEEE/CVF Winter Conference on Applications of Computer Vision}, pages 720--730, 2022.

\bibitem[Ferenc et~al.(2005)Ferenc, collaboration, et~al.]{ferenc2005magic}
Daniel Ferenc, MAGIC collaboration, et~al.
\newblock The magic gamma-ray observatory.
\newblock \emph{Nuclear Instruments and Methods in Physics Research Section A: Accelerators, Spectrometers, Detectors and Associated Equipment}, 553\penalty0 (1-2):\penalty0 274--281, 2005.

\bibitem[Garreau and von Luxburg(2020)]{garreau2020looking}
Damien Garreau and Ulrike von Luxburg.
\newblock Looking deeper into tabular lime.
\newblock \emph{arXiv preprint arXiv:2008.11092}, 2020.

\bibitem[Ghorbani et~al.(2019)Ghorbani, Abid, and Zou]{ghorbani2019interpretation}
Amirata Ghorbani, Abubakar Abid, and James Zou.
\newblock Interpretation of neural networks is fragile.
\newblock In \emph{Proceedings of the AAAI Conference on Artificial Intelligence}, volume~33, pages 3681--3688, 2019.

\bibitem[Gouk et~al.(2021)Gouk, Frank, Pfahringer, and Cree]{gouk2021regularisation}
Henry Gouk, Eibe Frank, Bernhard Pfahringer, and Michael~J Cree.
\newblock Regularisation of neural networks by enforcing lipschitz continuity.
\newblock \emph{Machine Learning}, 110\penalty0 (2):\penalty0 393--416, 2021.

\bibitem[Guidotti et~al.(2018)Guidotti, Monreale, Ruggieri, Turini, Giannotti, and Pedreschi]{guidotti2018survey}
Riccardo Guidotti, Anna Monreale, Salvatore Ruggieri, Franco Turini, Fosca Giannotti, and Dino Pedreschi.
\newblock A survey of methods for explaining black box models.
\newblock \emph{ACM computing surveys (CSUR)}, 51\penalty0 (5):\penalty0 1--42, 2018.

\bibitem[Hardt et~al.(2016)Hardt, Recht, and Singer]{hardt2016train}
Moritz Hardt, Ben Recht, and Yoram Singer.
\newblock Train faster, generalize better: Stability of stochastic gradient descent.
\newblock In \emph{International conference on machine learning}, pages 1225--1234. PMLR, 2016.

\bibitem[Hill et~al.(2022)Hill, Masoomi, Ghimire, Torop, and Dy]{hill2022explanation}
Davin Hill, Aria Masoomi, Sandesh Ghimire, Max Torop, and Jennifer Dy.
\newblock Explanation uncertainty with decision boundary awareness.
\newblock \emph{arXiv preprint arXiv:2210.02419}, 2022.

\bibitem[Holter et~al.(2018)Holter, Gomez, and Bertini]{holter2018fico}
Steffen Holter, Oscar Gomez, and Enrico Bertini.
\newblock Fico explainable machine learning challenge, 2018.

\bibitem[Jordan and Freiburger(2015)]{jordan2015effect}
Kareem~L Jordan and Tina~L Freiburger.
\newblock The effect of race/ethnicity on sentencing: Examining sentence type, jail length, and prison length.
\newblock \emph{Journal of Ethnicity in Criminal Justice}, 13\penalty0 (3):\penalty0 179--196, 2015.

\bibitem[Ju et~al.(2022)Ju, Zhang, Yang, Jiang, Liu, and Zhao]{ju2022logic}
Yiming Ju, Yuanzhe Zhang, Zhao Yang, Zhongtao Jiang, Kang Liu, and Jun Zhao.
\newblock Logic traps in evaluating attribution scores.
\newblock In \emph{Proceedings of the 60th Annual Meeting of the Association for Computational Linguistics (Volume 1: Long Papers)}, pages 5911--5922, 2022.

\bibitem[Krizhevsky et~al.(2009)Krizhevsky, Hinton, et~al.]{cifar10}
Alex Krizhevsky, Geoffrey Hinton, et~al.
\newblock Learning multiple layers of features from tiny images.
\newblock 2009.

\bibitem[Lakkaraju et~al.(2020)Lakkaraju, Arsov, and Bastani]{lakkaraju2020robust}
Himabindu Lakkaraju, Nino Arsov, and Osbert Bastani.
\newblock Robust and stable black box explanations.
\newblock In \emph{International Conference on Machine Learning}, pages 5628--5638. PMLR, 2020.

\bibitem[LeCun and Cortes(2010)]{mnist}
Yann LeCun and Corinna Cortes.
\newblock {MNIST} handwritten digit database.
\newblock 2010.
\newblock URL \url{http://yann.lecun.com/exdb/mnist/}.

\bibitem[Lei et~al.(2018)Lei, G’Sell, Rinaldo, Tibshirani, and Wasserman]{lei2018distribution}
Jing Lei, Max G’Sell, Alessandro Rinaldo, Ryan~J Tibshirani, and Larry Wasserman.
\newblock Distribution-free predictive inference for regression.
\newblock \emph{Journal of the American Statistical Association}, 113\penalty0 (523):\penalty0 1094--1111, 2018.

\bibitem[Li et~al.(2020)Li, Nagarajan, Plumb, and Talwalkar]{li2020learning}
Jeffrey Li, Vaishnavh Nagarajan, Gregory Plumb, and Ameet Talwalkar.
\newblock A learning theoretic perspective on local explainability.
\newblock \emph{arXiv preprint arXiv:2011.01205}, 2020.

\bibitem[Lundberg and Lee(2017)]{lundberg2017unified}
Scott~M Lundberg and Su-In Lee.
\newblock A unified approach to interpreting model predictions.
\newblock In \emph{Proceedings of the 31st international conference on neural information processing systems}, pages 4768--4777, 2017.

\bibitem[Lundberg et~al.(2018)Lundberg, Erion, and Lee]{treeshap}
Scott~M Lundberg, Gabriel~G Erion, and Su-In Lee.
\newblock Consistent individualized feature attribution for tree ensembles.
\newblock \emph{arXiv preprint arXiv:1802.03888}, 2018.

\bibitem[Mangal et~al.(2020)Mangal, Sarangmath, Nori, and Orso]{mangal2020probabilistic}
Ravi Mangal, Kartik Sarangmath, Aditya~V Nori, and Alessandro Orso.
\newblock Probabilistic lipschitz analysis of neural networks.
\newblock In \emph{International Static Analysis Symposium}, pages 274--309. Springer, 2020.

\bibitem[Masoomi et~al.(2020)Masoomi, Wu, Zhao, Wang, Castaldi, and Dy]{masoomi2020instance}
Aria Masoomi, Chieh Wu, Tingting Zhao, Zifeng Wang, Peter Castaldi, and Jennifer Dy.
\newblock Instance-wise feature grouping.
\newblock \emph{Advances in Neural Information Processing Systems}, 33, 2020.

\bibitem[Masoomi et~al.(2021)Masoomi, Hill, Xu, Hersh, Silverman, Castaldi, Ioannidis, and Dy]{masoomi2022explanations}
Aria Masoomi, Davin Hill, Zhonghui Xu, Craig~P. Hersh, Edwin~K. Silverman, Peter~J. Castaldi, Stratis Ioannidis, and Jennifer Dy.
\newblock Explanations of black-box models based on directional feature interactions.
\newblock In \emph{International Conference on Learning Representations}, 2021.

\bibitem[Nakkiran et~al.(2021)Nakkiran, Kaplun, Bansal, Yang, Barak, and Sutskever]{nakkiran2021deep}
Preetum Nakkiran, Gal Kaplun, Yamini Bansal, Tristan Yang, Boaz Barak, and Ilya Sutskever.
\newblock Deep double descent: Where bigger models and more data hurt.
\newblock \emph{Journal of Statistical Mechanics: Theory and Experiment}, 2021\penalty0 (12):\penalty0 124003, 2021.

\bibitem[Novak et~al.(2018)Novak, Bahri, Abolafia, Pennington, and Sohl-Dickstein]{novak2018sensitivity}
Roman Novak, Yasaman Bahri, Daniel~A Abolafia, Jeffrey Pennington, and Jascha Sohl-Dickstein.
\newblock Sensitivity and generalization in neural networks: an empirical study.
\newblock \emph{arXiv preprint arXiv:1802.08760}, 2018.

\bibitem[Petsiuk et~al.(2018)Petsiuk, Das, and Saenko]{petsiuk2018rise}
Vitali Petsiuk, Abir Das, and Kate Saenko.
\newblock Rise: Randomized input sampling for explanation of black-box models.
\newblock \emph{arXiv preprint arXiv:1806.07421}, 2018.

\bibitem[Ribeiro et~al.(2016)Ribeiro, Singh, and Guestrin]{ribeiro2016should}
Marco~Tulio Ribeiro, Sameer Singh, and Carlos Guestrin.
\newblock " why should i trust you?" explaining the predictions of any classifier.
\newblock In \emph{Proceedings of the 22nd ACM SIGKDD international conference on knowledge discovery and data mining}, pages 1135--1144, 2016.

\bibitem[Rieger and Hansen(2020)]{rieger2020simple}
Laura Rieger and Lars~Kai Hansen.
\newblock A simple defense against adversarial attacks on heatmap explanations.
\newblock In \emph{5th Annual Workshop on Human Interpretability in Machine Learning}, 2020.

\bibitem[Rosca et~al.(2020)Rosca, Weber, Gretton, and Mohamed]{pmlr-v137-rosca20a}
Mihaela Rosca, Theophane Weber, Arthur Gretton, and Shakir Mohamed.
\newblock A case for new neural network smoothness constraints.
\newblock In Jessica Zosa~Forde, Francisco Ruiz, Melanie~F. Pradier, and Aaron Schein, editors, \emph{Proceedings on "I Can't Believe It's Not Better!" at NeurIPS Workshops}, volume 137 of \emph{Proceedings of Machine Learning Research}, pages 21--32. PMLR, 12 Dec 2020.
\newblock URL \url{https://proceedings.mlr.press/v137/rosca20a.html}.

\bibitem[Schwab and Karlen(2019)]{schwab2019cxplain}
Patrick Schwab and Walter Karlen.
\newblock Cxplain: Causal explanations for model interpretation under uncertainty.
\newblock \emph{arXiv preprint arXiv:1910.12336}, 2019.

\bibitem[Smilkov et~al.(2017)Smilkov, Thorat, Kim, Vi{\'e}gas, and Wattenberg]{smilkov2017smoothgrad}
Daniel Smilkov, Nikhil Thorat, Been Kim, Fernanda Vi{\'e}gas, and Martin Wattenberg.
\newblock Smoothgrad: removing noise by adding noise.
\newblock \emph{arXiv preprint arXiv:1706.03825}, 2017.

\bibitem[Strobl et~al.(2008)Strobl, Boulesteix, Kneib, Augustin, and Zeileis]{strobl2008conditional}
Carolin Strobl, Anne-Laure Boulesteix, Thomas Kneib, Thomas Augustin, and Achim Zeileis.
\newblock Conditional variable importance for random forests.
\newblock \emph{BMC bioinformatics}, 9\penalty0 (1):\penalty0 1--11, 2008.

\bibitem[Tan and Tian(2023)]{tan2023robust}
Zeren Tan and Yang Tian.
\newblock Robust explanation for free or at the cost of faithfulness.
\newblock 2023.

\bibitem[Torop et~al.(2023)Torop, Masoomi, Hill, Kose, Ioannidis, and Dy]{torop2023smoothhess}
Max Torop, Aria Masoomi, Davin Hill, Kivanc Kose, Stratis Ioannidis, and Jennifer Dy.
\newblock Smoothhess: Re{LU} network feature interactions via stein's lemma.
\newblock In \emph{Thirty-seventh Conference on Neural Information Processing Systems}, 2023.
\newblock URL \url{https://openreview.net/forum?id=dwIeEhbaD0}.

\bibitem[Upadhyay et~al.(2021)Upadhyay, Joshi, and Lakkaraju]{upadhyay2021towards}
Sohini Upadhyay, Shalmali Joshi, and Himabindu Lakkaraju.
\newblock Towards robust and reliable algorithmic recourse.
\newblock \emph{Advances in Neural Information Processing Systems}, 34:\penalty0 16926--16937, 2021.

\bibitem[Virmaux and Scaman(2018)]{virmaux2018lipschitz}
Aladin Virmaux and Kevin Scaman.
\newblock Lipschitz regularity of deep neural networks: analysis and efficient estimation.
\newblock \emph{Advances in Neural Information Processing Systems}, 31, 2018.

\bibitem[Wang et~al.(2020)Wang, Wang, Ramkumar, Mardziel, Fredrikson, and Datta]{wang2020smoothed}
Zifan Wang, Haofan Wang, Shakul Ramkumar, Piotr Mardziel, Matt Fredrikson, and Anupam Datta.
\newblock Smoothed geometry for robust attribution.
\newblock \emph{Advances in neural information processing systems}, 33:\penalty0 13623--13634, 2020.

\bibitem[Yeh et~al.(2019)Yeh, Hsieh, Suggala, Inouye, and Ravikumar]{yeh2019fidelity}
Chih-Kuan Yeh, Cheng-Yu Hsieh, Arun Suggala, David~I Inouye, and Pradeep~K Ravikumar.
\newblock On the (in) fidelity and sensitivity of explanations.
\newblock \emph{Advances in Neural Information Processing Systems}, 32, 2019.

\bibitem[Yeh and Lien(2009)]{yeh2009comparisons}
I-Cheng Yeh and Che-hui Lien.
\newblock The comparisons of data mining techniques for the predictive accuracy of probability of default of credit card clients.
\newblock \emph{Expert systems with applications}, 36\penalty0 (2):\penalty0 2473--2480, 2009.

\bibitem[Yin et~al.(2021)Yin, Shi, Hsieh, and Chang]{yin2021faithfulness}
Fan Yin, Zhouxing Shi, Cho-Jui Hsieh, and Kai-Wei Chang.
\newblock On the faithfulness measurements for model interpretations.
\newblock \emph{arXiv preprint arXiv:2104.08782}, 2021.

\bibitem[Yoon et~al.(2018)Yoon, Jordon, and van~der Schaar]{yoon2018invase}
Jinsung Yoon, James Jordon, and Mihaela van~der Schaar.
\newblock Invase: Instance-wise variable selection using neural networks.
\newblock In \emph{International Conference on Learning Representations}, 2018.

\bibitem[Zhou et~al.(2022)Zhou, Ribeiro, and Shah]{zhou2022exsum}
Yilun Zhou, Marco~Tulio Ribeiro, and Julie Shah.
\newblock Exsum: From local explanations to model understanding.
\newblock \emph{arXiv preprint arXiv:2205.00130}, 2022.

\bibitem[Zintgraf et~al.(2017)Zintgraf, Cohen, Adel, and Welling]{zintgraf2017visualizing}
Luisa~M Zintgraf, Taco~S Cohen, Tameem Adel, and Max Welling.
\newblock Visualizing deep neural network decisions: Prediction difference analysis.
\newblock \emph{arXiv preprint arXiv:1702.04595}, 2017.

\end{thebibliography}
\bibliographystyle{plainnat}

% \subsubsection*{References}

% References follow the acknowledgements.  Use an unnumbered third level
% heading for the references section.  Please use the same font
% size for references as for the body of the paper---remember that
% references do not count against your page length total.

% \begin{thebibliography}{}
% \setlength{\itemindent}{-\leftmargin}
% \makeatletter\renewcommand{\@biblabel}[1]{}\makeatother
% \bibitem{} J.~Alspector, B.~Gupta, and R.~B.~Allen (1989).
%     \newblock Performance of a stochastic learning microchip.
%     \newblock In D. S. Touretzky (ed.),
%     \textit{Advances in Neural Information Processing Systems 1}, 748--760.
%     San Mateo, Calif.: Morgan Kaufmann.

% \bibitem{} F.~Rosenblatt (1962).
%     \newblock \textit{Principles of Neurodynamics.}
%     \newblock Washington, D.C.: Spartan Books.

% \bibitem{} G.~Tesauro (1989).
%     \newblock Neurogammon wins computer Olympiad.
%     \newblock \textit{Neural Computation} \textbf{1}(3):321--323.
% \end{thebibliography}

%%%%%%%%%%%%%%%%%%%%%%%%%%%%%%%%%%%%%%%%%%%%%%%%%%%%%%%%%%%%
\clearpage
\section*{Checklist}

% % %%% BEGIN INSTRUCTIONS %%%
% The checklist follows the references. For each question, choose your answer from the three possible options: Yes, No, Not Applicable.  You are encouraged to include a justification to your answer, either by referencing the appropriate section of your paper or providing a brief inline description (1-2 sentences). 
% Please do not modify the questions.  Note that the Checklist section does not count towards the page limit. Not including the checklist in the first submission won't result in desk rejection, although in such case we will ask you to upload it during the author response period and include it in camera ready (if accepted).

% \textbf{In your paper, please delete this instructions block and only keep the Checklist section heading above along with the questions/answers below.}
% %%% END INSTRUCTIONS %%%

 \begin{enumerate}

 \item For all models and algorithms presented, check if you include:
 \begin{enumerate}
   \item A clear description of the mathematical setting, assumptions, algorithm, and/or model. [Yes/No/\textbf{Not Applicable}] This paper does not present any models or algorithms.
   \item An analysis of the properties and complexity (time, space, sample size) of any algorithm. [Yes/No/\textbf{Not Applicable}] This paper does not present any models or algorithms.
   \item (Optional) Anonymized source code, with specification of all dependencies, including external libraries. [Yes/No/\textbf{Not Applicable}] This paper does not present any models or algorithms. Code for experiments is provided in Supplementary Materials.
 \end{enumerate}

 \item For any theoretical claim, check if you include:
 \begin{enumerate}
   \item Statements of the full set of assumptions of all theoretical results. [\textbf{Yes}/No/Not Applicable] All theorems in Section 4 specify all needed assumptions.
   \item Complete proofs of all theoretical results. [\textbf{Yes}/No/Not Applicable]  Proof for Theorem 1 is in main text, all other proofs are provided in the Appendix.
   \item Clear explanations of any assumptions. [\textbf{Yes}/No/Not Applicable]     
 \end{enumerate}

 \item For all figures and tables that present empirical results, check if you include:
 \begin{enumerate}
   \item The code, data, and instructions needed to reproduce the main experimental results (either in the supplemental material or as a URL). [\textbf{Yes}/No/Not Applicable] Code for experiments is provided as part of supplementary materials.
   \item All the training details (e.g., data splits, hyperparameters, how they were chosen). [Yes\st{/No/Not Applicable}] Training details are provided in the experiment section and in the Appendix.
         \item A clear definition of the specific measure or statistics and error bars (e.g., with respect to the random seed after running experiments multiple times). [\textbf{Yes}/No/Not Applicable] Error bars are explained where used.
         \item A description of the computing infrastructure used. (e.g., type of GPUs, internal cluster, or cloud provider). [Yes/\textbf{No}/Not Applicable] We did not keep track of this and don't consider it to be important to report for this work.
 \end{enumerate}

 \item If you are using existing assets (e.g., code, data, models) or curating/releasing new assets, check if you include:
 \begin{enumerate}
   \item Citations of the creator If your work uses existing assets. [\textbf{Yes}/No/Not Applicable] Citations are included for all datasets used.
   \item The license information of the assets, if applicable. [Yes/No/\textbf{Not Applicable}]
   \item New assets either in the supplemental material or as a URL, if applicable. [Yes/No/\textbf{Not Applicable}]
   \item Information about consent from data providers/curators. [Yes/No/\textbf{Not Applicable}]
   \item Discussion of sensible content if applicable, e.g., personally identifiable information or offensive content. [Yes/No/\textbf{Not Applicable}]
 \end{enumerate}

 \item If you used crowdsourcing or conducted research with human subjects, check if you include:
 \begin{enumerate}
   \item The full text of instructions given to participants and screenshots. [Yes/No/\textbf{Not Applicable}]
   \item Descriptions of potential participant risks, with links to Institutional Review Board (IRB) approvals if applicable. [Yes/No/\textbf{Not Applicable}]
   \item The estimated hourly wage paid to participants and the total amount spent on participant compensation. [Yes/No/\textbf{Not Applicable}]
 \end{enumerate}

 \end{enumerate}

\clearpage
\appendix
\onecolumn
\section{DETAILED PROOFS}
\label{app:proof}
We include the detailed proofs for Lemma~\ref{lem1}, Theorems~\ref{thm2} and \ref{thm3} here.

\subsection{Lemma \ref{lem1}}

\begin{proof}
Let us assume,

\begin{equation*}
    p_k = \mathbb{P}[\mathbb{N}_k], s.t. \mathbb{N}_k=\{x \mid x \in \mathbb{R}^d, ||x||_0 = k, x_i \neq 0\}
\end{equation*}

and let $\hat{\mathbb{L}}$ be the set of points that violate Lipschitzness, then assume,

\begin{equation*}
    \gamma_k = \mathbb{P}[\hat{\mathbb{L}} \mid \mathbb{N}_k]
\end{equation*}

given that $\alpha$ is the probability of the set of points that violate Lipschitzness across $\mathcal{D}$, we can use Bayes' rule to write,

\begin{equation*}
    \alpha = \mathbb{P}[\hat{\mathbb{L}}] =  \sum_{k=1}^d p_k \gamma_k
\end{equation*}

If we consider the case where the sets $\mathbb{N}_k$ are finite, each $\mathbb{N}_k$ can be mapped to a set $\mathbb{N}'_k$ of cardinality,
\begin{equation*}
    |\mathbb{N}'_k| = \sum_{b=0}^{d-k} {d-k \choose b} |\mathbb{N}_k| = 2^{d-k}|\mathbb{N}_k|
\end{equation*}
In more general terms, the probability of $\mathbb{N}'_k$ can be written as,
\begin{equation*}
    p'_k = \mathbb{P}[\mathbb{N}'_k] = \frac{2^{d-k}p_k}{\sum_{j=1}^d 2^{d-j} p_j} = \frac{2^{-k}p_k}{\sum_{j=1}^d 2^{-j} p_j}
\end{equation*}
Let us define $\beta$ as the proportion of points in \textit{all} $\mathbb{N}'_k$ that also violate Lipschitzness in their unmasked form. This leads us to the following equation for $\beta$
\small\begin{equation*}
\beta = \frac{\sum_{k=1}^d 2^{-k} p_k \gamma_k}{\sum_{j=1}^d 2^{-j} p_j}    
\end{equation*}
\normalsize 
The worse case $\beta$ would then be obtained by considering a maximization over $\gamma_k$,
\small\begin{equation}
% \begin{split}
\beta^* = \max_{\gamma_1, \ldots ,\gamma_d} \frac{\sum_{k=1}^d 2^{-k} p_k \gamma_k}{\sum_{j=1}^d 2^{-j} p_j},\sum_{i=1}^d p_i \gamma_i = \alpha, 0 \leq \alpha \leq 1 ,0 \leq \gamma_i \leq 1, \forall i=1,\ldots,d      
% \end{split}
\end{equation}
\normalsize
This constrained optimization problem can be solved by assigning $\gamma_k=1$ for the largest $p_k$ until the budget $\alpha$ is exhausted where only a fractional value of $\gamma$ can be assigned, and $0$ for the remaining values of $k$. This $\beta^* \geq \alpha$ in general. In the specific case where $p_k \rightarrow 0$ for $k=2,\ldots,d$, when compared to $p_1$ (i.e. where the probability of sampling a point from $\mathcal{D}$ such that any of the values are \emph{exactly} $0$ is very small compared to the probability of sampling points with all non-zero values which would generally be the case for sampling real data), $\beta^* \rightarrow \alpha$
\end{proof}

\subsection{Theorem~\ref{thm3}}
\begin{proof}
 By considering another point $x'$ such that $d_p(x,x') \leq r$ and \eqref{eq:removal_exp} we get,
    
    \small\begin{equation}
        d_p(\phi(x_i), \phi(x'_i)) = d_p(f(x) - f(x \odot z_{-i}), f(x') - f(x' \odot z_{-i}))
    \end{equation}
    
    using the fact that $d_p(x,y) = ||x-y||_p$ where  $||.||_p$ is the $p$-norm, the RHS gives us,
    
    \small\begin{equation}
        d_p(\phi_i(x), \phi_i(x')) = ||f(x) - f(x \odot z_{-i}) - f(x') + f(x' \odot z_{-i})||_p
    \end{equation}
    
    using triangular inequality,
    \small\begin{equation}
        d_p(\phi_i(x), \phi_i(x')) \leq ||f(x) - f(x')||_p + ||f(x' \odot z_{-i}) - f(x \odot z_{-i}) ||_p
    \end{equation}
    
    w.l.o.g assuming the first term on the right is bigger than the second term
    
    \small\begin{equation}
        d_p(\phi_i(x), \phi_i(x')) \leq 2||f(x) - f(x')||_p = 2d_p(f(x), f(x'))
    \end{equation}
    
    using the fact that $f$ is probabilistic Lipschitz get us,
    
    \small\begin{equation}
        \mathbb{P}[d_p(\phi_i(x), \phi_i(x')) \leq 2Ld_p(x,x')] \geq 1 - \alpha
    \end{equation}
    
    to conclude the proof note that $d_p(\phi(x), \phi(x')) \leq \sqrt[p]{d} * \max_i d_p(\phi_i(x),\phi_i(x'))$, which gives us,
    
    \small\begin{equation}
        \mathbb{P}[d_p(\phi(x), \phi(x')) \leq 2 \sqrt[p]{d}L \cdot d_p(x,x')] \geq 1-\alpha
    \end{equation}

\end{proof}

\subsection{Theorem~\ref{thm2}}
\begin{proof}

    Given input $x$ and another input $x'$ s.t. $d(x,x') \leq r$, using \eqref{eq:rise} we can write
   
   \small\begin{equation}
    \begin{split}
    d_p(\phi_i(x),\phi_i(x')) &= d_p(\mathbb{E}_{p(z|z_i=1)}[f(x \odot z)] ,\mathbb{E}_p(z|z_i=1)[f(x' \odot z)]) \\
    &=||\mathbb{E}_{p(z|z_i=1)}[f(x \odot z)] -\mathbb{E}_p(z|z_i=1)[f(x' \odot z)]||_p \\
    &=||\mathbb{E}_{p(z|z_i=1)}[f(x \odot z) - f(x' \odot z)]||_p 
    \end{split}
   \end{equation}
   
   Using Jensen's inequality on R.H.S,
   
   \small\begin{equation}
       d_p(\phi_i(x),\phi_i(x')) \leq \mathbb{E}_{p(z|z_i=1)}[||f(x \odot z) - f(x' \odot z)||_p] 
   \end{equation}
   
   Using the fact that $E[f] \leq \max f$,
   
    \small\begin{equation}
        \begin{split}
            d_p(\phi_i(x),\phi_i(x')) &\leq \max_z ||f(x \odot z) - f(x' \odot z)||_p \\
            &=\max_z d_p(f(x \odot z),f(x' \odot z))
        \end{split}
       \label{eq:risep1}
   \end{equation}
   
   Using the fact that $f$ is deterministically Lipschitz with some constant $L \geq 0$, and $d_p(x \odot z, x' \odot z) \leq d_p(x,x'), \forall z$. Then using the definition of probabilistic Lipschitz with $\alpha=0$ we get,
   
   \small\begin{equation}
       \mathbb{P}(\max_z d_p(f(x \odot z),f(x' \odot z)) \leq L*d(x,x') \geq 1
   \end{equation}
   
   Using this in \eqref{eq:risep1} gives us,
   
   \small\begin{equation}
       \mathbb{P}[ d_p(\phi_i(x),\phi_i(x'))\leq L*d(x,x')] \geq 1
   \label{eq:risef}
   \end{equation}
   
   Note that \eqref{eq:risef} is true for each feature $i \in \{1, ..., d\}$. To conclude the proof note that $d_p(\phi(x), \phi(x') \leq \sqrt[p]{d} * \max_i d_p(\phi_i(x),\phi_i(x'))$. Utilizing this with \eqref{eq:risef} leads us to
   
   \small\begin{equation}
       \mathbb{P}[ d_p(\phi(x),\phi(x') \leq \sqrt[p]{d}L \cdot d_p(x,x')] \geq 1
       \label{eq:riseall}
   \end{equation}
   
   Since $\mathbb{P}[d_p(\phi(x),\phi(x') \leq \sqrt[p]{d}L \cdot d_p(x,x')]$ defines $A_{\lambda,r}$ for $\lambda \geq \sqrt[p]{d}L$, this concludes the proof.
\end{proof}
\section{DATASET DETAILS}
\label{app:datasets}
\begin{itemize}
    \item \textbf{Orange-skin}: The input data is again generated from a 10-dimensional standard Gaussian distribution. The ground truth class probabilities are proportional to $\exp\{ \sum_{i=1}^4 X_i^2 - 4 \}$. In this case the first 4 features are important globally for \emph{all} data points.
    
    \item \textbf{Nonlinear-additive}: Similar to \textit{Orange-skin} dataset except the ground trugh class probabilities are proportional to $\exp \{-100\sin{2X_1} + 2|X_2| + X_3 + \exp\{-X_4\} \}$, and therefore each of the 4 important features for prediction are nonlinearly related to the prediction itself.
    
    \item \textbf{Switch}: This simulated dataset is specifically for instancewise feature explanations. For the input data feature $X_1$ is generated by a mixture of Gaussian distributions centered at $\pm 3$. If $X_1$ is generated from the Gaussian distribution centered at $+3$, $X_2$ to $X_5$ are used to generate the prediction probabilities according to the \emph{Orange skin} model. Otherwise $X_6$ to $X_9$ are used to generate the prediction probabilities according to the \emph{Nonlinear-additive} model.
    
    \item \textbf{Rice} \citep{cinar2019classification}:This dataset consists of $3810$ samples of rice grains of two different varieties (\emph{Cammeo} and \emph{Osmancik}). 7 morphological features are provided for each sample.
    
    \item \textbf{Telescope}\citep{ferenc2005magic}: This dataset consists of $19000+$ Monte-Carlo generated samples to simulate registration of high energy gamma particles in a ground-based atmospheric Cherenkov gamma telescope using the imaging technique. Each sample is labelled as either background or gamma signal and consists of 10 features.
    
\end{itemize}

\section{TRAINING DETAILS} \label{app:training_details}

Training splits and hyperparameter choices have relatively little effect on our experiments. Regardless, the details used in results shown are provided here for completeness:

\begin{itemize}
    \item \textbf{Train/Test Split:} For all synthetic datasets we use $10^6$ training points and $10^3$ test points. The neural networks classifiers were trained with a batch size of $1000$ for $2$ epochs. While SVM was trained with default parameters used in \url{https://scikit-learn.org/stable/modules/generated/sklearn.svm.SVC.html}.
    
    For Telescope and Rice datasets test set sizes of $5\%$ and $33\%$ were used, with a batch size of $32$ trained for $100$ epochs. SVM was again trained with default parameters.
    
    \item \textbf{radius $r$:} For all experiments we used radius equal to the median of pairwise distance. This is standard practice and also allows for a big enough $r$ where we can sample enough points to provide empirical estimates.

    \item \textbf{Classifier details: } We train the following four classifiers; \textbf{2layer} : A two-layer MLP with ReLU activations. For simulated datasets each layer has $200$ neurons, while for the four real datasets we use $32$ neurons in each layer. \textbf{4layer}: A four-layer MLP with ReLU activations, with the same number of neurons per layer as \emph{2layer}. \textbf{linear}: A linear classifier. \textbf{svm}: A support vector machine with Gaussian kernel
\end{itemize}
\begin{table}[t]
    {\centering
    % This represents the average tightness of the lower bound.
    \setlength{\belowcaptionskip}{-20pt}
    \scriptsize
    % \small
    \begin{center}
    \setlength\tabcolsep{3.2pt}
        \begin{tabular}{lcccccccc}
        \toprule
        & \multicolumn{2}{c}{2layer}
        & \multicolumn{2}{c}{4layer}
        & \multicolumn{2}{c}{linear}
        & \multicolumn{2}{c}{svm}\\
        \cmidrule(lr){2-3}
        \cmidrule(lr){4-5}
        \cmidrule(lr){6-7}
        \cmidrule(lr){8-9}
            Datasets      
            & Train
            & Test
            & Train
            & Test   
            & Train
            & Test    
            & Train
            & Test
            \\           
        \midrule          	
            	CIFAR10
            	& .435
            	& .416
            	& .457
            	& .443
            	& .385
            	& .375
            	& .526
            	& .512
                    \\  	
            	MNIST
            	& .939
            	& .940
            	& .968
            	& .969
            	& .907
            	& .910
            	& .983
            	& .980
            	\\     
        \bottomrule  
        \end{tabular}
    \end{center}
    }
    \caption{Train and Test accuracy for different classifiers used in the experiments in Section~5.2.} 
    \label{table:acc_class}
    \end{table}

\begin{table}[t]
    {\centering
    % This represents the average tightness of the lower bound.
    \setlength{\belowcaptionskip}{-20pt}
    \scriptsize
    % \small
    \begin{center}
    \setlength\tabcolsep{3.2pt}
        \begin{tabular}{lcccccc}
        \toprule
        & \multicolumn{2}{c}{Regularized High}
        & \multicolumn{2}{c}{Regularized Low}
        & \multicolumn{2}{c}{Not Regularized}\\
        \cmidrule(lr){2-3}
        \cmidrule(lr){4-5}
        \cmidrule(lr){6-7}
            Datasets      
            & Train
            & Test
            & Train
            & Test   
            & Train
            & Test
            \\           
        \midrule          	
            	CIFAR10
            	& .338
            	& .328
            	& .518
            	& .491
            	& .529
            	& .497
                    \\  	
            	MNIST
            	& .788
            	& .809
            	& .951
            	& .969
            	& .990
            	& .976
            	\\     
        \bottomrule  
        \end{tabular}
    \end{center}
    }
    \caption{Train and Test accuracy for different classifiers used in the experiments in Section~5.1.} 
    \label{table:acc_reg}
    \end{table}
\section{ADDITIONAL RESULTS}
\label{app:results}
Table~\ref{table:sup_tab} shows the normalized AUC for the estimated explainer astuteness and the predicted AUC based on the predicted lower bound curve. As expected the predicted AUC lower bounds the estimated AUC.

\begin{figure}[t]
    \centering
        \includegraphics[width=0.75\columnwidth]{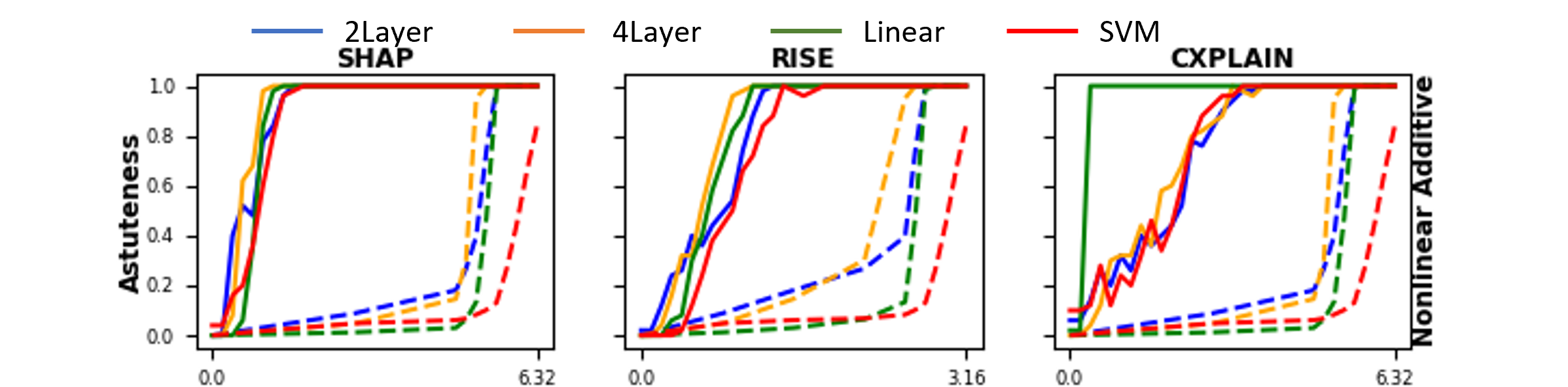}
    \includegraphics[width=0.75\columnwidth]{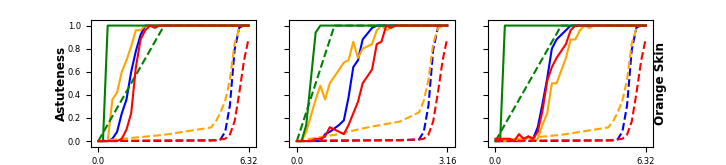}
    \includegraphics[width=0.75\columnwidth]{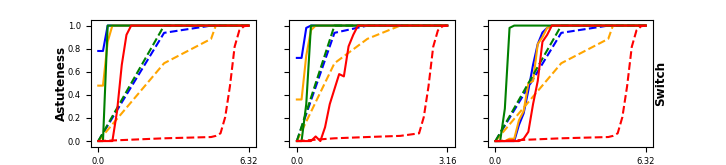}
    \includegraphics[width=0.75\columnwidth]{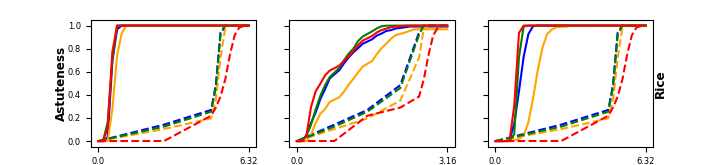}
    \includegraphics[width=0.75\columnwidth]{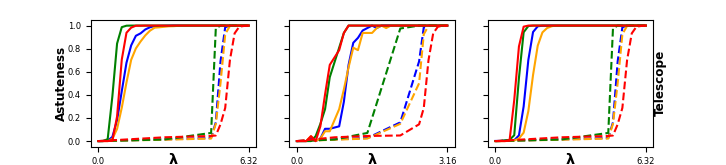}
    \includegraphics[width=0.75\columnwidth]{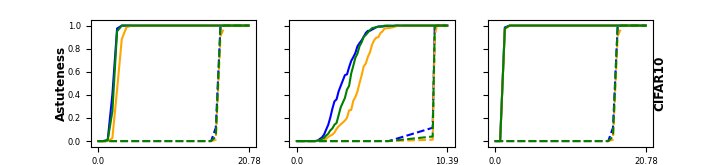}
    \includegraphics[width=0.75\columnwidth]{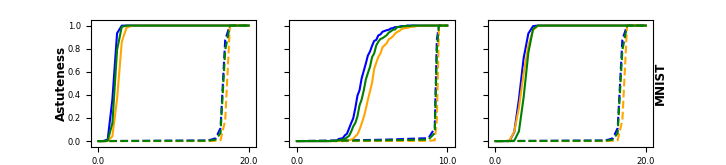}
    \caption{This figure experimentally shows the implication of our theoretical results. It corresponds to the AUC values shown in Table~\ref{table:auc}. Given each combination of dataset, classifier and explainer we observe that the estimated explainer astuteness for SHAP, RISE and CXPLAIN is lower bounded by the astuteness predicted by our theoretical results given a value of $\lambda$. The predicted lower bound is depicted by dashed lines, while solid lines depict the actual estimate of explainer astuteness.}
    \label{fig:sup_experiments}
\end{figure}

Figure~\ref{fig:sup_experiments} shows the same plots as shown in Figure~\ref{fig:experiments2} but includes all datasets.

\begin{table*}
    {\centering
\caption{\textbf{Observed AUC and (Predicted AUC)}. The observed AUC is lower bounded by the predicted AUC and so the observed AUC should always be higher than the predicted AUC. The AUC values are normalized between 0 and 1.}
\label{table:sup_tab}
\scriptsize
    \begin{center}
    \setlength\tabcolsep{3.0pt}
        \begin{tabular}{c|c|c|c|c|c|c|c|c|c|c|c|c|c|c|c|c|}
        & \multicolumn{4}{c|}{2layer}
        & \multicolumn{4}{c}{4layer}
        & \multicolumn{4}{|c}{linear}
        & \multicolumn{4}{|c|}{svm}\\
        \midrule
            \textbf{Datasets}      
            & \textbf{SHAP}
            & \textbf{RISE}
            & \textbf{CXP}
            & \textbf{(LB)}      
            & \textbf{SHAP}
            & \textbf{RISE}
            & \textbf{CXP}
            & \textbf{(LB)}      
            & \textbf{SHAP}
            & \textbf{RISE}
            & \textbf{CXP}
            & \textbf{(LB)}       
            & \textbf{SHAP}
            & \textbf{RISE}
            & \textbf{CXP}
            & \textbf{(LB)}  
            \\           
        \midrule
            	OS
            	& .954
            	& .847
            	& .920
            	& (.369)
            	& .969
            	& .896
            	& .906
            	& (.480)
            	& .994
            	& .967
            	& .994
            	& (.950)
            	& .945
            	& .813
            	& .917
            	& (.184)
            	\\   
            	NA
            	& .978
            	& .909
            	& .936
            	& (.618)
            	& .981
            	& .926
            	& .940
            	& (.696)
            	& .972
            	& .912
            	& .994
            	& (.520)
            	& .971
            	& .883
            	& .937
            	& (.229)
            	\\             	
            	Switch
            	& .998
            	& .996
            	& .948
            	& (.945)
            	& .996
            	& .988
            	& .948
            	& (.909)
            	& .994
            	& .978
            	& .988
            	& (.950)
            	& .969
            	& .885
            	& .936
            	& (.412)
            	\\            	
            	Rice
            	& .962
            	& .886
            	& .974
            	& (.803)
            	& .932
            	& .824
            	& .932
            	& (.793)
            	& .968
            	& .901
            	& .962
            	& (.800)
            	& .981
            	& .906
            	& .970
            	& (.715)
            	\\             	
            	Telescope
            	& .962
            	& .863
            	& .954
            	& (.637)
            	& .955
            	& .863
            	& .944
            	& (.610)
            	& .980
            	& .906
            	& .967
            	& (.756)
            	& .969
            	& .909
            	& .972
            	& (.467)
            	\\              	
            	CIFAR10
            	& .994
            	& .898
            	& .998
            	& (.661)
            	& .992
            	& .866
            	& .998
            	& (.642)
            	& .994
            	& .888
            	& .998
            	& (.653)
            	& .996
            	& .990
            	& .919
            	& (.661)
            	\\               	
            	MNIST
            	& .994
            	& .853
            	& .998
            	& (.553)
            	& .992
            	& .826
            	& .988
            	& (.469)
            	& .993
            	& .842
            	& .986
            	& (.538)
            	& -
            	& -
            	& -
            	& (-)
            	\\   
        \midrule  
        \end{tabular}
    \end{center}
    }

    \end{table*}

\begin{figure}[t]
    \centering
    \includegraphics[width=1\columnwidth]{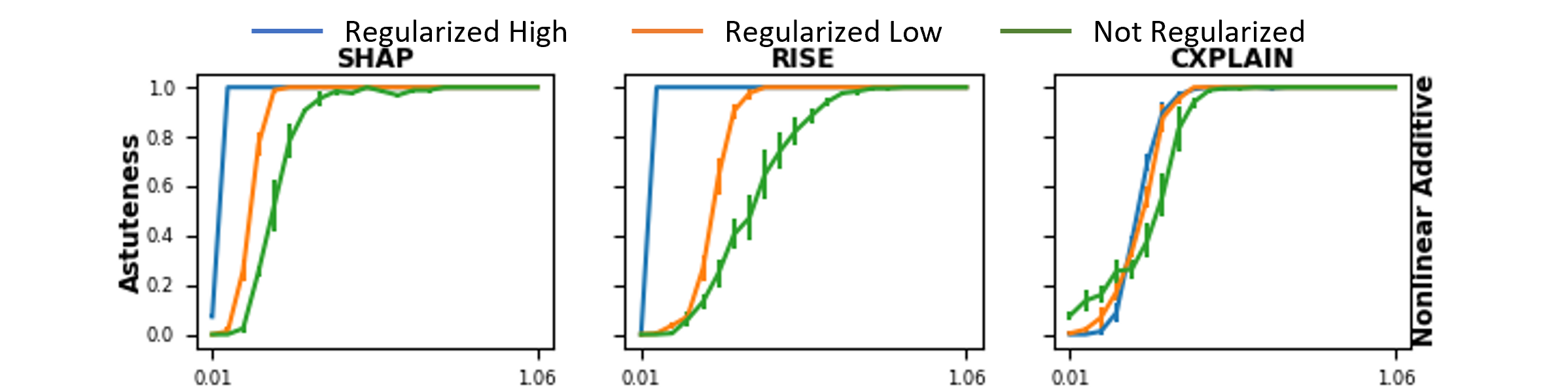}
    \includegraphics[width=1\columnwidth]{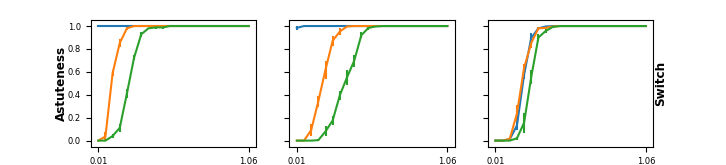}
    \includegraphics[width=1\columnwidth]{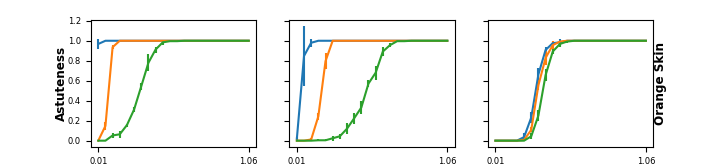}
    \includegraphics[width=1\columnwidth]{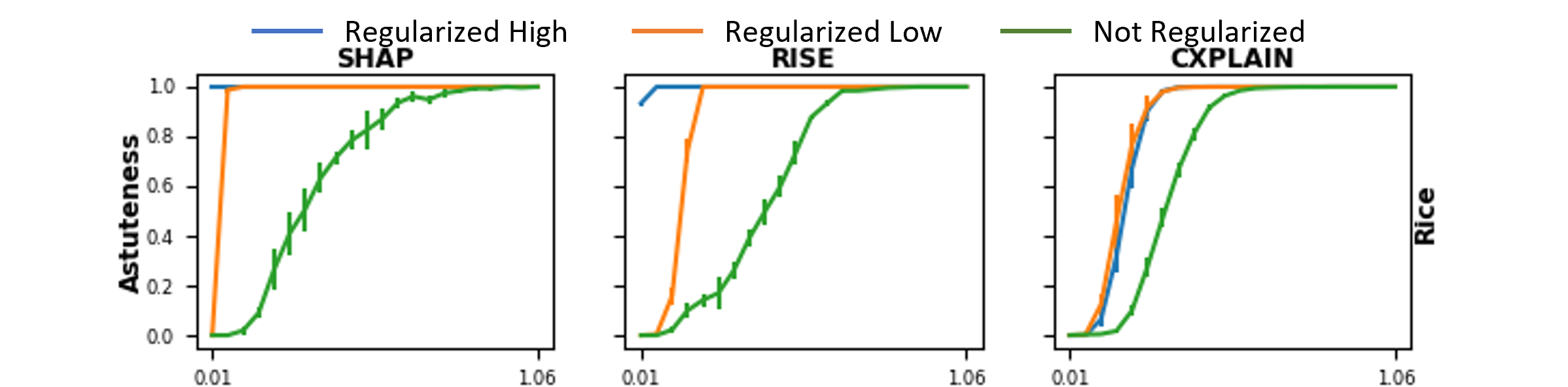}
    \includegraphics[width=1\columnwidth]{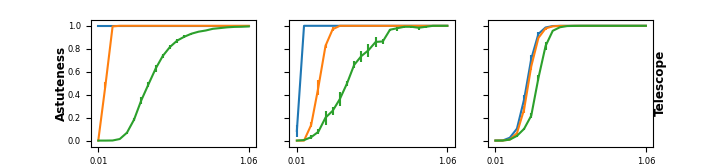}
    \caption{Regularizing the Lipschitness of a neural network during training results in higher astuteness for the same value of $\lambda$. Higher regularization results in lower Lipschitz constant \citep{gouk2021regularisation}. Astuteness reaches $1$ for smaller values of $\lambda$ with Lipschitz regularized training, as expected from our theorems. The errorbars represent results across 5 runs to account for randomness in training.}  
    \label{fig:experiments2_sup}
\end{figure}

\end{document}